%% file: main.tex
\documentclass{techreport}

\title{DynaSolver: A Neural Operator for Autoregressive 4D Flow Prediction with Dynamic Boundaries}
\shorttitle{DynaSolver}   
\author{
Rui Qian\affil{1}$^{*}$ \quad
Shibo Dai\affil{1}$^{*}$ \quad
Mengjun Lin\affil{1}$^{*}$ \quad
Yuk-Kit Hou\affil{2} \\
\textbf{Saining Zhang}\affil{3}$^{\dagger}$ \quad
\textbf{Bohan Li}\affil{4} \quad
\textbf{Junhao Chen}\affil{1} \quad
\textbf{Hao Zhao}\affil{1} \quad
\textbf{Chaojian Li}\affil{5}$^{\ddagger}$
}
\affiliation{
    \affil{1}Tsinghua University \quad
    \affil{2}Peking University \quad
    \affil{3}Nanyang Technological University \\
    \affil{4}Shanghai Jiao Tong University \quad
    \affil{5}The Hong Kong University of Science and Technology
}
\metadata{Project Page}{\href{https://made-gpt.github.io/DynaSolver-ProjectPage/}{DynaSolver-ProjectPage}}

\begin{document}

\begin{abstract}
Neural operators provide efficient surrogates for computational fluid dynamics (CFD), but most are designed for static geometries and are not equipped to model unsteady flows induced by continuously moving boundaries. We study \emph{dynamic-boundary 4D CFD prediction} for articulated moving geometries and introduce \textbf{RoboCFD-4D}, a temporally resolved robotic CFD benchmark spanning diverse morphologies and flow conditions.
To address the computational and autoregressive challenges of this setting, we propose \textbf{DynaSolver}, a neural operator that explicitly conditions on boundary evolution and causal flow history. 
DynaSolver employs an efficient physical-token formulation that amortizes point-to-token projection across the operator stack, substantially reducing repeated point-space spatio-temporal computation while retaining accurate point-level reconstruction. 
We further introduce Replay-Augmented Parallel Training, which exposes the model to autoregressively generated histories while preserving parallel dense supervision, narrowing the gap between ground-truth-conditioned training and autoregressive rollout.
Experiments on RoboCFD-4D show that DynaSolver achieves the best full-rollout accuracy among a broad range of learned CFD surrogates, with its advantage increasing as rollout proceeds, indicating substantially reduced autoregressive error accumulation. 
\end{abstract}

\maketitle

\tableofcontents

\input{sections/00_introduction}
\input{sections/01_related_work}

\input{sections/02_method}
\input{sections/03_experiments}

\input{sections/04_conclusion}

\clearpage
\input{sections/05_appendix}

\bibliographystyle{plainnat}
\bibliography{references}

\end{document}

%% file: sections/00_introduction.tex
\section{Introduction}
\label{sec:introduction}

Computational fluid dynamics (CFD) is a fundamental tool for analyzing complex flow phenomena, and recent learning-based surrogates, particularly neural operators, have substantially reduced the computational cost of high-fidelity flow prediction. Most existing progress~\citep{fno2021,Liu2026PhysGTO,mspt2025,linearno2026,abupt2025,Wu2024Transolver}, however, has focused on static aerodynamic configurations such as airfoils~\citep{DBLP:conf/nips/BonnetMCG22}, vehicles~\citep{elrefaie2025drivaernet,DBLP:conf/nips/ElrefaieMDA24,abupt2025}, and fixed industrial components~\citep{li2023gino,deng2024geometry}, where the geometry remains unchanged during prediction. Meanwhile, robotic systems are increasingly operating in high-speed, outdoor, and fluid-interactive environments~\citep{paolino2025learning,loquercio2021learning}, where aerodynamic effects can directly influence locomotion stability and control~\citep{jones2022physics,lale2024falcon}. Efficient prediction of the surrounding flow field is therefore becoming increasingly relevant to robotic systems~\citep{du2021underwater,hu2019chainqueen}. 
Unlike conventional aerodynamic objects, robots exhibit articulated geometry and motion, whose joint modeling has received increasing attention in 3D reconstruction~\citep{shen2026gaussianart,gao2025partrm,zhang2024s3o}. In CFD, such time-varying geometry acts as a continuously moving boundary, violating the static-boundary assumption underlying most existing CFD surrogates~\citep{Wu2024Transolver,li2023gino,Cao2023BSMSGNN,DBLP:conf/icml/Belbute-PeresEK20}.

Flow prediction around a moving robot cannot be reduced to independently applying a static CFD surrogate to each instantaneous geometry. Under prescribed motion, the evolving boundary continuously drives the surrounding fluid~\citep{peskin2002immersed,sotiropoulos2014immersed}, while previously developed flow structures persist over time. More generally, finite or partially observed dynamical representations can exhibit memory effects~\citep{parish2017non,ayed2019learning}, so similar robot configurations may correspond to different flow states depending on their preceding evolution. We therefore formulate \emph{dynamic-boundary 4D CFD prediction}, where both the three-dimensional boundary and surrounding flow evolve over time.

\begin{center}
    \centering
    \captionsetup{type=figure}
\includegraphics[width=1.0\textwidth,height=1\textheight,keepaspectratio]{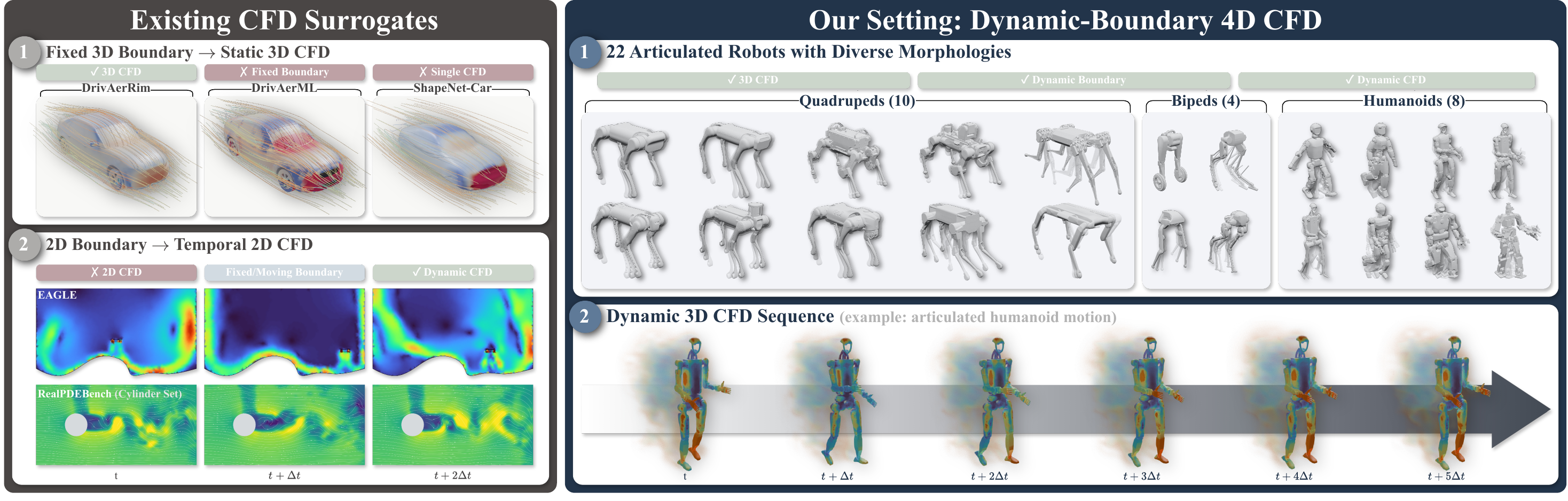}
    \caption{\textbf{Existing CFD Settings versus Dynamic-Boundary 4D CFD.} Existing CFD surrogates typically model either static three-dimensional geometries or temporally evolving two-dimensional flows. In contrast, RoboCFD-4D requires joint modeling of articulated three-dimensional boundary motion and the resulting time-evolving flow field across diverse robot morphologies.}
    \label{fig:teaser}
\end{center}%

Learning such dynamics introduces three challenges beyond conventional static-boundary CFD prediction. First, the model must represent both boundary evolution and causal flow context rather than treating each geometry independently. 
Second, extending neural operators to temporally resolved three-dimensional CFD substantially increases computational cost, since repeated point-space processing compounds across spatial resolution, temporal context, and network depth.
Third, autoregressive rollout recursively feeds predicted states back into the model, creating a mismatch with ground-truth-conditioned training and causing prediction errors to accumulate over time~\citep{list2025differentiability,brandstetter2022message,kochkov2021machine}. Dynamic-boundary CFD therefore requires a model that captures causal boundary-flow evolution, performs spatio-temporal reasoning efficiently, and remains robust under closed-loop rollout.

\begin{table*}[htbp]
\centering
\caption{
Comparison of representative CFD learning datasets across spatial dimensionality, temporal resolution, and boundary dynamics. “Dim.” denotes spatial and temporal dimensionality, where 3D+t indicates temporally resolved three-dimensional CFD.
}
\label{tab:cfd_dataset_comparison}

\resizebox{\textwidth}{!}{
\begin{tabular}{lccccccc}
\toprule[1.4pt]
\textbf{Dataset} &
\textbf{Domain} &
\textbf{Dim.} &
\textbf{Boundary Dynamics} &
\textbf{Sequences} & 
\textbf{CFD States} \\
\midrule

AirfRANS~\citep{DBLP:conf/nips/BonnetMCG22}
& Airfoil aerodynamics
& 2D
& Static solid
& N/A
& 1,000 \\

CFDBench~\citep{DBLP:journals/corr/abs-2310-05963}
& General fluid dynamics
& 2D+t
& Static solid
& 739
& 301,671 \\

MegaFlow2D~\citep{DBLP:conf/cpsweek/XuGM23}
& External / internal flow
& 2D+t
& Static solid
& $\sim$3,000
& $>$2M \\

RealPDEBench~\citep{hu2026realpdebench}
& real scenarios
& 2D+t 
& Static/Moving solid 
& 736
& $\sim$2.6M\\

Eagle~\citep{janny2023eagle}
& Drone
& 2D+t
& Moving Source, Static solid 
& 1184
& $\sim$1M \\

FlowBench~\citep{tali2025flowbench}
& Complex-geometry flow
& 2D+t
& Static solid
& 1,150
& 278,300 \\

ShapeNet-Car~\citep{DBLP:journals/tog/UmetaniB18}
& Automotive aerodynamics
& 3D
& Static solid
& N/A
& 889 \\

NASA CRM~\citep{Bekemeyer2025IntroductionOA}
& Aircraft aerodynamics
& 3D
& Static solid
& N/A
& 121 \\

AhmedML~\citep{DBLP:journals/corr/abs-2407-20801}
& Bluff-body aerodynamics
& 3D
& Static solid
& N/A
& 500 \\

DrivAerML~\citep{DBLP:journals/corr/abs-2408-11969}
& Automotive aerodynamics
& 3D
& Static solid
& N/A
& 500 \\

DrivAerRim~\citep{xia2026drivaerrim}
& Automotive aerodynamics
& 3D
& Static solid
& N/A
& 904 \\

DrivAerNet~\citep{elrefaie2025drivaernet}
& Automotive aerodynamics
& 3D
& Static solid
& N/A
& 4,000 \\

DrivAerNet++~\citep{DBLP:conf/nips/ElrefaieMDA24}
& Automotive aerodynamics
& 3D
& Static solid
& N/A
& $\sim$8,000 \\

\midrule

\textbf{RoboCFD-4D} (Ours)
& \textbf{Robot aerodynamics}
& \textbf{3D+t}
& \textbf{Articulated moving solid}
& \textbf{352}
& \textbf{10560} \\

\bottomrule[1.4pt]
\end{tabular}
}
\end{table*}

To enable systematic study of this problem, we construct \textbf{RoboCFD-4D}, a temporally resolved three-dimensional CFD benchmark based on articulated robotic motion. As summarized in Table~\ref{tab:cfd_dataset_comparison}, existing CFD learning datasets predominantly consider static three-dimensional geometries~\citep{elrefaie2025drivaernet,DBLP:conf/nips/ElrefaieMDA24,DBLP:conf/nips/BonnetMCG22,abupt2025} or temporally evolving two-dimensional flows~\citep{janny2023eagle,hu2026realpdebench}, whereas RoboCFD-4D contains 352 sequences and 10,560 CFD frames with articulated moving three-dimensional boundaries.
Building on this benchmark, we propose \textbf{DynaSolver} based on a simple principle: dynamic-boundary CFD should be modeled as causal evolution between physical flow states under prescribed boundary motion, rather than as independent geometry-to-flow predictions. DynaSolver therefore combines explicit boundary evolution and causal flow history with an efficient physical-token temporal operator, enabling compact spatio-temporal reasoning without repeatedly returning to the full point representation. We further introduce \emph{Replay-Augmented Parallel Training}, which augments clean training histories with autoregressively generated states while retaining parallel dense supervision, improving rollout robustness without the cost of sequential student optimization.

Our main contributions are summarized as follows:
\begin{itemize}
\item We formulate dynamic-boundary 4D CFD prediction for temporally evolving three-dimensional flow fields and introduce \textbf{RoboCFD-4D}, containing 352 sequences and 10,560 time-resolved CFD frames with articulated moving boundaries.

\item We propose \textbf{DynaSolver}, a neural operator that jointly models explicit boundary evolution and causal flow dynamics. 
Its physical-token temporal operator eliminates layer-wise point--token
alternation, performing repeated spatio-temporal reasoning in compact token
space with only a single slice/deslice pair.

\item We introduce \textbf{Replay-Augmented Parallel Training}, which combines autoregressively generated histories with parallel dense supervision, exposing the model to rollout states without the sequential student optimization required by dense autoregressive training.

\end{itemize}

%% file: sections/01_related_work.tex
\section{Related Work}
\label{sec:related}

\paragraph{CFD datasets with evolving dynamics.}
Existing benchmarks such as PDEBench and The Well~\citep{takamoto2022pdebench,ohana2024well} cover diverse spatiotemporal physical systems. For CFD, AirfRANS provides steady 2D airfoil flows, while DrivAerNet and DrivAerNet++ offer static 3D vehicle aerodynamics~\citep{DBLP:conf/nips/BonnetMCG22,elrefaie2025drivaernet,DBLP:conf/nips/ElrefaieMDA24}. FlowBench further covers steady and transient flows over complex 2D/3D geometries~\citep{tali2025flowbench}. Moving-boundary settings remain more limited: ~\citep{paolino2025learning} study articulated humanoid aerodynamics under a quasi-steady assumption, while EAGLE and RealPDEBench consider time-resolved 2D flows with moving sources or solids~\citep{janny2023eagle,hu2026realpdebench}, and pitching-airfoil simulations involve prescribed motion of simpler geometries~\citep{BALDAN2024109345}. In contrast, RoboCFD-4D provides transient 3D CFD trajectories around complex articulated bodies with continuously evolving boundaries.

\paragraph{Learning-based CFD surrogates.}
Neural operators provide an efficient framework for learning mappings between physical fields. DeepONet~\citep{lu2021deeponet} uses branch trunk factorization, while Fourier-based approaches such as FNO~\citep{fno2021} and its variants model global interactions through spectral representations~\citep{lu2021deeponet,fno2021,rahman2023uno,wen2022ufno,gupta2021multiwavelet,tran2023ffno,wu2023lsm}. Geometry-aware operators, including GNO, Geo-FNO, and GINO, extend operator learning to irregular domains~\citep{Li2020NeuralOperator,Li2023GeoFNO,li2023gino}, while attention-based methods such as GNOT and Transolver provide direct global interactions on irregular discretizations~\citep{hao2023gnot,Wu2024Transolver}. Mesh-based simulators such as GNS, MeshGraphNet, and BSMS-GNN instead propagate local interactions through graph message passing~\citep{sanchez2020learning,Pfaff2021MeshGraphNet,Cao2023BSMSGNN}. 
Physics-informed approaches further incorporate physical constraints and governing equations into learning objectives~\citep{PINN,Li2021PINO,zhang2024fluidmotion}. Building on these advances, DynaSolver focuses on efficient spatio-temporal operator learning under continuously evolving articulated boundaries.

\paragraph{Autoregressive training for neural simulation.}
Autoregressive neural simulators are prone to error accumulation because model predictions are recursively reused as future inputs. Noise injection approximates this rollout corruption by perturbing ground-truth states during training~\citep{sanchez2020learning}, while unrolled training exposes the model directly to its own predictions at the cost of sequential rollout and increased optimization expense. The pushforward strategy reduces gradient cost by detaching intermediate predictions while retaining autoregressive state generation~\citep{brandstetter2022message}. 
Subsequent work and systematic benchmarks further examine autoregressive training strategies, rollout stability, and long-horizon prediction errors~\citep{list2025differentiability,koehler2024apebench}.
Our Replay-Augmented Parallel Training instead decouples autoregressive history generation from student optimization, combining model-generated replay states with parallel dense supervision.

%% file: sections/02_method.tex
\section{Method}
\label{sec:method}

We study dynamic-boundary 4D CFD prediction for articulated robots under prescribed motion, predicting the evolution of the surrounding flow from the current state and known boundary motion. After formalizing the problem in Sec.~\ref{sec:problem} and introducing the pipeline in Sec.~\ref{sec:overview}, we present the token-space temporal operator in Sec.~\ref{sec:efficient_operator} and Replay-Augmented Parallel Training in Sec.~\ref{sec:replay_training}.


\begin{figure}[t]
\includegraphics[width=0.98\linewidth]{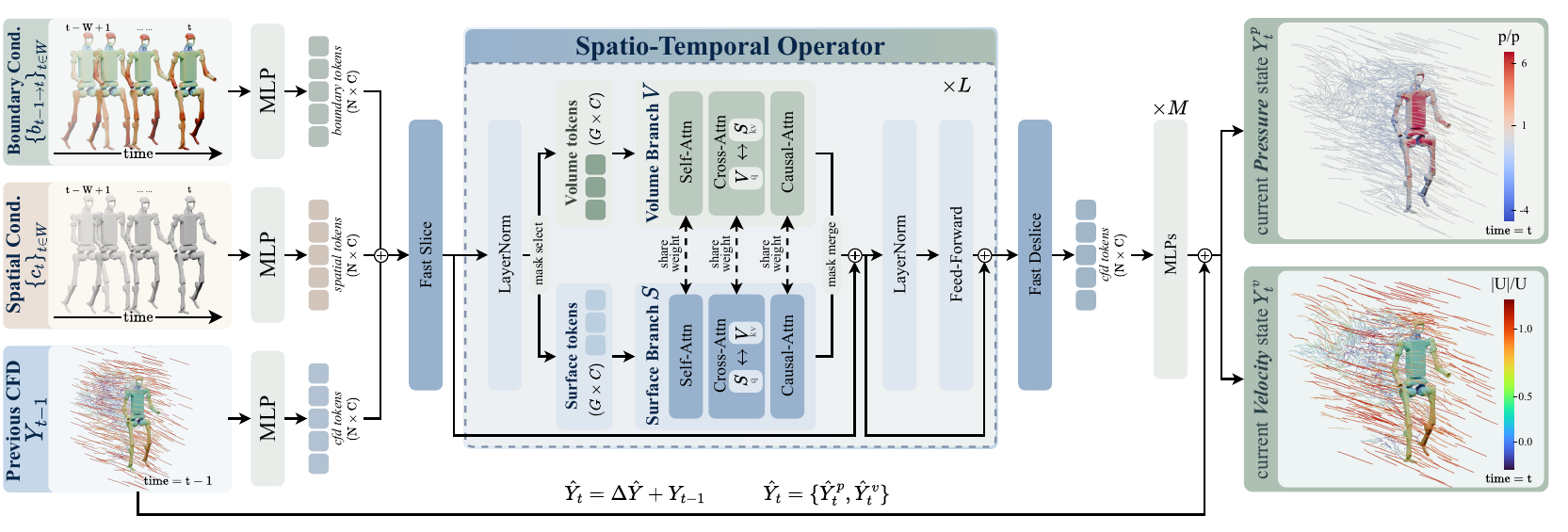}
\centering
\caption{\textbf{Overview of DynaSolver.}
Given the spatial condition, boundary evolution, and previous CFD states over a causal window, DynaSolver lifts and fuses the point-wise inputs before mapping them into separate surface and volume physical tokens. A single Fast Slice/Deslice pair encloses the spatio-temporal operator stack, where self-attention, cross-attention, and causal attention jointly model spatial and temporal flow interactions. The reconstructed point features are decoded into a residual flow field and added to the previous CFD state to predict the current velocity and pressure.}
\label{fig:pipeline}
\end{figure}

\subsection{Problem Setup}
\label{sec:problem}

We consider one-way coupled transient incompressible flow around an articulated robot undergoing prescribed motion, where the moving robot boundary drives the surrounding fluid dynamics while the robot motion itself is not affected by the predicted flow. 
The surrogate operates on a fixed Eulerian model grid $\mathcal{X}=\{\mathbf{x}_i\}_{i=1}^{N}$. A set of tracked material points follows the robot surface and is projected onto $\mathcal{X}$ to define time-dependent surface-associated masks and geometric and kinematic features. Grid nodes occupied by the moving robot are excluded through a time-dependent validity mask $M_t^{\mathrm{v}}$. The CFD state is represented on the fixed grid as $Y_t=[u,v,w,p]_t\in\mathbb{R}^{N\times4}$, where $(u,v,w)$ and $p$ denote velocity and kinematic pressure, respectively.

Under a prescribed moving boundary, advancing the flow from $t-1$ to $t$ requires not only the previous CFD state but also the time-dependent boundary configuration and kinematics~\citep{peskin2002immersed,sotiropoulos2014immersed}.
We therefore condition the prediction on a spatial condition $c_t\in\mathbb{R}^{N\times11}$ describing the configuration at time $t$, and a boundary-evolution feature $b_{t-1\rightarrow t}\in\mathbb{R}^{N\times8}$ describing the boundary motion over the transition, yielding the one-step mapping $(c_t,b_{t-1\rightarrow t},Y_{t-1})\mapsto Y_t$.

Although the complete CFD state under prescribed boundary evolution admits a local Markov transition, projection onto a finite or partially observed representation can introduce memory effects~\citep{parish2017non,ayed2019learning}. 
Our surrogate operates precisely in such a setting, 
with time-varying boundary features and validity masks defined on a fixed Eulerian representation, 
and a single observed state is therefore not guaranteed to be Markov-sufficient. We consequently augment the transition with a short causal history. Defining the transition-aligned input $q_i=(c_i,b_{i-1\rightarrow i},Y_{i-1})$, we collect at most $W$ consecutive transitions as
\begin{equation}
    \mathcal{Q}_t
    =
    \left\{q_i\right\}_{i=\max(1,t-W+1)}^{t}.
    \label{eq:causal_window}
\end{equation}

We then learn a temporal operator $\mathcal{F}_{\theta}$ over this causal context and predict the flow evolution in residual form,
\begin{equation}
    \widehat{Y}_t
    =
    Y_{t-1}
    +
    \mathcal{F}_{\theta}(\mathcal{Q}_t).
    \label{eq:temporal_operator}
\end{equation}
Starting from the given initial state $Y_0$, the causal context grows until $W$ transitions are available and then proceeds with a fixed-length sliding window.


\subsection{Method Overview}
\label{sec:overview}
Dynamic 4D CFD introduces two challenges beyond spatial flow prediction. 
First, temporal reasoning over multiple flow states substantially amplifies the cost of point-space computation. 
Second, efficient parallel training relies on ground-truth histories, whereas autoregressive inference recursively conditions on model predictions, creating a train--inference distribution gap. DynaSolver addresses these challenges with an efficient token-space temporal operator and Replay-Augmented Parallel Training, respectively.

Given the causal transition window $\mathcal{Q}_t$, the three input modalities of each transition are first lifted into a shared latent space,
\begin{equation}
    H_i=\phi_c(c_i)+\phi_b(b_{i-1\rightarrow i})+\phi_y(Y_{i-1}),
    \label{eq:input_lifting}
\end{equation}
where $\phi_c$, $\phi_b$, and $\phi_y$ are two-layer MLPs. As illustrated in Fig.~\ref{fig:pipeline}, DynaSolver slices the fused point features once into separate surface and volume physical tokens, processes them with a stack of spatio-temporal operator blocks, and deslices them only before final point reconstruction. A lightweight MLP decodes the reconstructed point features into the residual field $\Delta Y_t$, which is added to $Y_{t-1}$ to obtain $\widehat{Y}_t$. During training, a frozen replay model generates autoregressive histories, while the current model retains parallel, densely supervised optimization through Replay-Augmented Parallel Training.


\subsection{Efficient Token-Space Temporal Operator}
\label{sec:efficient_operator}

The original Transolver~\citep{Wu2024Transolver} alternates between point and physical-token representations within each operator layer. Since the slicing weights are recomputed from the updated point features, the physical-token representation can evolve as the features are progressively refined. Carrying this layer-wise point--token alternation into temporal CFD, however, becomes costly because each transition contains thousands of surface and volume queries. 
We therefore amortize the physical tokenization across the operator stack: point-wise features are sliced once into a compact physical-token representation, all spatio-temporal reasoning is performed in this space, and the resulting features are desliced only at the output, as illustrated in Fig.~\ref{fig:pipeline}.

\textbf{Token-space operator.} We use the fast slicing operation of Transolver3~\citep{DBLP:journals/corr/abs-2602-04940} to compress the lifted point features into separate surface and volume physical-token streams,
\begin{equation}
    Z_i^{\mathrm{s}}, Z_i^{\mathrm{v}}
    =
    \mathrm{FastSlice}(H_i),
    \qquad
    Z_i^{\mathrm{s}}, Z_i^{\mathrm{v}}
    \in \mathbb{R}^{G\times D}.
    \label{eq:fast_slice}
\end{equation}
The resulting tokens are processed by $L$ spatio-temporal operator blocks. Each block applies intra-domain self-attention, bidirectional cross-attention between the surface and volume branches, causal temporal attention, and a token-space feed-forward network. Causal temporal attention~\citep{ashish2017transformer} uses a block-causal mask over transitions, allowing each transition to aggregate information only from the current and preceding transitions.

\textbf{Lightweight point reconstruction.} After the final operator block, the surface and volume streams are mapped back to their corresponding query points through a single fast deslicing operation,
\begin{equation}
    \widetilde H_i
    =
    \mathrm{FastDeslice}
    \left(
        Z_i^{\mathrm{s}},
        Z_i^{\mathrm{v}}
    \right).
    \label{eq:fast_deslice}
\end{equation}
Because this restructuring removes the repeated point-wise refinement of the original backbone, a linear prediction head is insufficient to recover the point-level flow field. We therefore use a lightweight MLP decoder $\phi_{\mathrm{dec}}$,
\begin{equation}
    \Delta Y_i
    =
    \phi_{\mathrm{dec}}(\widetilde H_i),
    \label{eq:flow_decoder}
\end{equation}
and add the residual to the previous flow state following Eq.~\ref{eq:temporal_operator}. 
The lightweight decoder compensates for the reduced point-level expressiveness introduced by the token-space restructuring, allowing the computational benefits of compact token processing to be realized without sacrificing prediction accuracy. As shown in Sec.~\ref{sec:ablation}, it recovers the performance degradation of the token-only variant with only modest additional overhead.


\begin{figure}[t]
\includegraphics[width=0.98\linewidth]{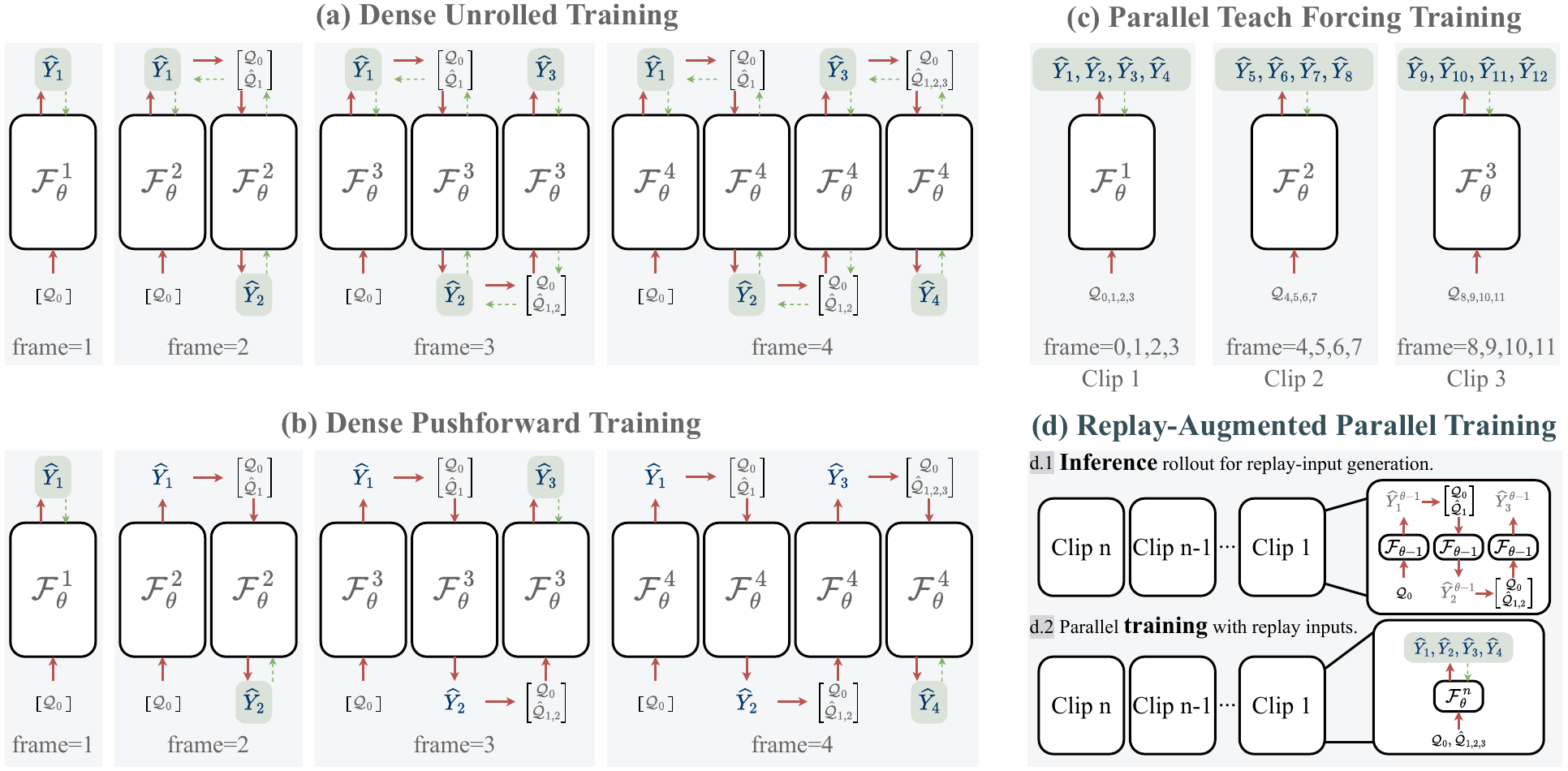}
\centering
\caption{\textbf{Comparison of training strategies for autoregressive flow prediction.}
Dense Unrolled Training (a) and Dense Pushforward Training (b) construct autoregressive histories sequentially, whereas Parallel Teacher-Forcing Training (c) predicts all transitions in a clip simultaneously from ground-truth histories. Our Replay-Augmented Parallel Training (d) decouples autoregressive history generation from student optimization: a frozen model first rolls out replay inputs (d.1), which are then used for parallel dense supervision of the current model (d.2). Red solid arrows \textcolor{red}{\(\rightarrow\)} denote feedforward computation, while green dashed arrows \textcolor{green!60!black}{\(\dashrightarrow\)} denote backpropagation.}
\label{fig:training_strategy}
\end{figure}

\subsection{Replay-Augmented Parallel Training}
\label{sec:replay_training}
Autoregressive flow prediction requires balancing rollout robustness with training efficiency. As illustrated in Fig.~\ref{fig:training_strategy}, \emph{Dense Unrolled Training}~\citep{list2025differentiability} exposes the model to its own predictions by sequentially unrolling the trajectory, but repeatedly evaluates highly overlapping temporal contexts and incurs substantial training cost. \emph{Dense Pushforward Training}~\citep{brandstetter2022message} reduces the backward cost by detaching preceding rollout steps, yet still requires sequential rollout construction. In contrast, \emph{Parallel Training}~\citep{ashish2017transformer} exploits block-causal attention to predict and supervise all transitions in a clip with a single forward pass, but conditions only on ground-truth histories and therefore suffers from a train--inference distribution gap. To combine autoregressive exposure with efficient parallel optimization, we introduce \emph{Replay-Augmented Parallel Training}: a frozen model first performs inference rollout to generate replay inputs (Fig.~\ref{fig:training_strategy}d.1), after which the current model is optimized in parallel using the replay-augmented clips (Fig.~\ref{fig:training_strategy}d.2).

\textbf{Replay generation.} We partition each trajectory into non-overlapping transition clips of length $W$. At epoch $e>1$, the previous-epoch model is frozen as the replay model, $\bar{\theta}^{(e)}=\theta^{(e-1)}$. For a fraction $\rho$ of the clips, the frozen model starts from the ground-truth initial state and rolls out autoregressively using the same causal-window mechanism as inference:
\begin{equation}
    \widetilde{Y}_i = widetilde{Y}_{i-1} + \mathcal{F}_{\bar{\theta}^{(e)}}
    \left(
        \widetilde{\mathcal{Q}}_i
    \right),
    \label{eq:replay_generation}
\end{equation}
where $\widetilde{\mathcal{Q}}_i$ follows the same growing-prefix and sliding-window construction as $\mathcal{Q}_i$, with ground-truth flow histories replaced by recursively generated replay states. Replay generation is detached from gradient computation, and the replay model remains fixed within each epoch.

\textbf{Parallel training and objective.} Unlike dense autoregressive training, the generated rollout is used only to construct input histories. As highlighted in Fig.~\ref{fig:training_strategy}, replay clips and the remaining clean ground-truth clips are then processed identically by the current model: all transition positions are predicted and supervised simultaneously in a single block-causal forward pass, while the targets remain ground-truth CFD states. For the supervised transition set $\mathcal{T}$, we optimize
\begin{equation}
    \mathcal{L}
    =
    \frac{1}{|\mathcal{T}|}
    \sum_{i\in\mathcal{T}}
    \frac{
        \sum_{n=1}^{N}
        m_{i,n}
        \left\|
            \widehat{Y}_{i,n}-Y_{i,n}
        \right\|_2^2
    }{
        \sum_{n=1}^{N} m_{i,n}
    },
    \label{eq:training_loss}
\end{equation}
where $m_{i,n}$ denotes the validity mask. Mixing replay and clean histories exposes the model to realistic model-induced errors while retaining a clean-state anchor, and the frozen previous-epoch model keeps the replay distribution fixed within each epoch.

%% file: sections/03_experiments.tex
\section{Experiments}
\label{sec:experiments}

We evaluate DynaSolver on RoboCFD-4D in terms of prediction accuracy, rollout stability, and efficiency, followed by controlled ablations on the model design and training strategy. Detailed training configurations and evaluation protocols are provided in the appendix (Sec.~\ref{sec:experimental_setup})

\begin{table*}[t]
\centering
\setlength{\tabcolsep}{3.5pt}
\renewcommand{\arraystretch}{1.15}

\caption{\textbf{Quantitative comparison across rollout horizons on RoboCFD-4D Base.}
We report the relative $L_2$ errors (\%) of velocity and pressure at selected rollout steps, together with their full-rollout averages and the overall error. The best results among learned methods are highlighted in
\colorbox[HTML]{DEF3E6}{\strut\textbf{bold}},
while the second-best are
\colorbox[HTML]{ECF8F1}{\strut\underline{underlined}}.
\textcolor{gray}{Persistence} is excluded from ranking.}
\label{tab:rollout_comparison}

\resizebox{0.98\textwidth}{!}{%
\begin{tabular}{l|cccccc|cccccc|ccc}
\toprule[1.4pt]

\multicolumn{1}{l|}{\multirow{2}{*}{\textbf{Method}}}
& \multicolumn{6}{c|}{\textbf{Per-frame Velocity} $\downarrow$}
& \multicolumn{6}{c|}{\textbf{Per-frame Pressure} $\downarrow$}
& \multicolumn{1}{c}{\multirow{2}{*}{\textbf{Avg. Vel.} $\downarrow$}}
& \multicolumn{1}{c}{\multirow{2}{*}{\textbf{Avg. Pre.} $\downarrow$}}
& \multicolumn{1}{c}{\multirow{2}{*}{\textbf{Overall} $\downarrow$}}
\\

& 1 & 6 & 12 & 18 & 24 & 29
& 1 & 6 & 12 & 18 & 24 & 29
& & & \\
\midrule

GNOT~\citep{hao2023gnot}
               & 11.38 & 21.14 & 28.47 & 34.58 & 40.01 & 43.99
               & 34.08 & 72.89 & 93.49 & 98.66 & 102.21 & 99.18
               & 30.56 & 87.84 & 31.55 \\

GAOT~\citep{NEURIPS2025_e45a448d}
               & 14.46 & 19.54 & 23.58 & 26.93 & 29.66 & 31.19
               & 40.96 & 57.70 & 66.52 & 75.18 & 84.08 & 90.67
               & 24.60 & 70.25 & 25.32 \\

CNO ~\citep{DBLP:conf/nips/RaonicMRRBAMB23}
               & 9.32 & 19.36 & 24.27 & 28.49 & 35.25 & 47.04
               & 34.24 & 77.09 & 87.29 & 98.14 & 108.89 & 120.71
               & 27.12 & 91.52 & 28.28 \\

P3D  ~\citep{DBLP:journals/corr/abs-2509-10186}
               & 9.32 & 17.07 & 22.62 & 26.70 & 30.44 & 32.36
               & 32.55 & 62.66 & 76.03 & 85.25 & 96.86 & 102.08
               & 23.72 & 78.82 & 24.61 \\

MGN~\citep{Pfaff2021MeshGraphNet}
               & 9.49 & 19.70 & 23.69 & 25.88 & 27.66 & 28.55
               & 29.32 & 65.02 & 79.25 & 86.55 & 96.78 & 99.86
               & 23.48 & 79.44 & 24.51 \\

DPOT~\citep{hao2024dpot}
               & 13.66 & 22.65 & 26.70 & 32.77 & 39.39 & 44.98
               & 42.45 & 78.12 & 103.77 & 133.25 & 166.67 & 187.97
               & 30.30 & 121.10 & 32.08 \\

AB-UPT  ~\citep{abupt2025}
               & 14.75 & 19.60 & 23.36 & 25.68 & 27.92 & 29.38
               & 42.29 & 77.47 & 91.30 & 96.30 & 97.54 & 96.23
               & 23.83 & 86.84 & 25.03 \\

UNO ~\citep{rahman2023uno}
               & 11.71 & 18.17 & 22.82 & 26.91 & 34.53 & 62.31
               & 34.13 & 58.36 & 71.43 & 81.76 & 104.69 & 156.99
               & 27.32 & 81.55 & 28.40 \\

U-Net ~\citep{DBLP:journals/corr/RonnebergerFB15}
               & \cellcolor[HTML]{DEF3E6}\textbf{7.48} & 16.73 & 20.99 & 23.30 & 25.05 & 26.08
               & 27.90 & 61.64 & 78.41 & 85.41 & 96.04 & 95.75
               & 20.81 & 77.99 & 21.86 \\

MSPT~\citep{mspt2025}
               & 9.28 & 19.15 & 23.18 & 25.89 & 28.14 & 29.94
               & \cellcolor[HTML]{ECF8F1}\underline{27.70} & 61.18 & 73.90 & 86.65 & 101.71 & 110.88
               & 23.42 & 79.14 & 24.39 \\

$\text{Transformer}_{\mathrm{galerkin}}$~\citep{Cao2021Gal}
               & 11.89 & 19.49 & 23.07 & 28.43 & 35.67 & 42.74
               & 38.61 & 69.13 & 96.66 & 115.08 & 136.97 & 168.67
               & 26.86 & 104.64 & 28.11 \\

DilResNet ~\citep{DBLP:journals/corr/abs-2112-15275}
               & 8.55 & 17.99 & 22.08 & 23.90 & 25.04 & 25.30
               & 29.00 & 61.87 & 78.00 & 83.80 & 86.84 & 84.99
               & 21.44 & 74.56 & 22.30 \\

PhysGTO~\citep{Liu2026PhysGTO}
               & 9.41 & 18.51 & 21.85 & 23.68 & 25.69 & 26.97
               & 30.52 & 60.97 & 70.12 & 78.40 & 89.10 & 92.79
               & 21.81 & 72.96 & 22.99 \\

GNS~\citep{sanchez2020learning}
               & 9.67 & 19.74 & 24.31 & 27.72 & 30.70 & 32.74
               & 29.69 & 61.92 & 78.24 & 89.28 & 99.89 & 106.83
               & 24.97 & 80.91 & 26.09 \\

DeepONet~\citep{lu2021deeponet}
               & 16.68 & 21.74 & 26.20 & 28.05 & 28.29 & 28.67
               & 60.07 & 73.64 & 86.37 & 91.34 & 92.49 & 90.63
               & 25.51 & 84.65 & 26.38 \\

GraphViT~\citep{janny2023eagle}
               & \cellcolor[HTML]{ECF8F1}\underline{8.33} & 17.05 & 20.41 & 22.18 & 23.61 & 24.44
               & 28.28 & 57.98 & 69.36 & 71.16 & 75.61 & 76.20
               & 20.15 & 66.05 & 21.00 \\

BSMS ~\citep{Cao2023BSMSGNN}
               & 8.44 & 17.30 & 21.20 & 24.12 & 26.66 & 28.85
               & \cellcolor[HTML]{DEF3E6}\textbf{27.01} & 54.20 & 61.57 & 68.21 & 77.84 & 81.24
               & 21.78 & 63.92 & 22.51 \\

FNO~\citep{fno2021}
               & 10.70 & 18.04 & 22.07 & 26.54 & 33.95 & 42.81
               & 31.13 & \cellcolor[HTML]{DEF3E6}\textbf{52.02} & 60.85 & 66.83 & \cellcolor[HTML]{ECF8F1}\underline{71.30} & 76.06
               & 25.49 & 61.69 & 26.08 \\

LinearNO ~\citep{linearno2026}
               & 10.26 & 16.81 & 19.74 & 21.66 & 22.92 & 23.84
               & 29.14 & \cellcolor[HTML]{ECF8F1}\underline{52.05} & \cellcolor[HTML]{ECF8F1}\underline{60.81} & \cellcolor[HTML]{ECF8F1}\underline{65.82} & 72.07 & \cellcolor[HTML]{ECF8F1}\underline{72.55}
               & 19.82 & \cellcolor[HTML]{ECF8F1}\underline{61.00} & 20.58 \\

MWT ~\citep{DBLP:journals/corr/abs-2109-13459}
               & 9.55 & 16.50 & \cellcolor[HTML]{ECF8F1}\underline{18.67} & \cellcolor[HTML]{ECF8F1}\underline{20.26} & \cellcolor[HTML]{ECF8F1}\underline{22.03} & \cellcolor[HTML]{ECF8F1}\underline{23.35}
               & 32.05 & 57.19 & 70.58 & 83.28 & 97.58 & 106.52
               & \cellcolor[HTML]{ECF8F1}\underline{18.97} & 76.64 & 20.11 \\
   
Transolver~\citep{Wu2024Transolver}
               & 10.19 & \cellcolor[HTML]{DEF3E6}\textbf{16.24} & 19.10 & 20.73 & 22.38 & 23.45
               & 29.77 & 54.27 & 61.51 & 66.99 & 74.53 & 75.64
               & 19.23 & 62.76 & \cellcolor[HTML]{ECF8F1}\underline{19.99} \\

\midrule

\textcolor{gray}{Persistence}
               & \textcolor{gray}{14.14} & \textcolor{gray}{20.78} & \textcolor{gray}{19.80} & \textcolor{gray}{20.42} & \textcolor{gray}{20.92} & \textcolor{gray}{21.35}
               & \textcolor{gray}{40.17} & \textcolor{gray}{60.19} & \textcolor{gray}{54.16} & \textcolor{gray}{54.63} & \textcolor{gray}{58.75} & \textcolor{gray}{58.21}
               & \textcolor{gray}{20.14} & \textcolor{gray}{56.09} & \textcolor{gray}{20.95} \\

\textbf{DynaSolver} (Ours)
               & 11.61 & \cellcolor[HTML]{ECF8F1}\underline{16.27} & \cellcolor[HTML]{DEF3E6}\textbf{17.37} & \cellcolor[HTML]{DEF3E6}\textbf{17.86} & \cellcolor[HTML]{DEF3E6}\textbf{18.06} & \cellcolor[HTML]{DEF3E6}\textbf{18.36}
               & 34.02 & 53.19 & \cellcolor[HTML]{DEF3E6}\textbf{57.48} & \cellcolor[HTML]{DEF3E6}\textbf{59.17} & \cellcolor[HTML]{DEF3E6}\textbf{60.37} & \cellcolor[HTML]{DEF3E6}\textbf{61.61}
               & \cellcolor[HTML]{DEF3E6}\textbf{17.07} & \cellcolor[HTML]{DEF3E6}\textbf{56.35} & \cellcolor[HTML]{DEF3E6}\textbf{17.71} \\

\bottomrule[1.4pt]
\end{tabular}%
}
\end{table*}

\begin{figure*}[htbp]
    \centering
    \captionsetup[subfigure]{skip=1pt}

    \begin{subfigure}{0.32\textwidth}
        \centering
        \includegraphics[width=\linewidth]{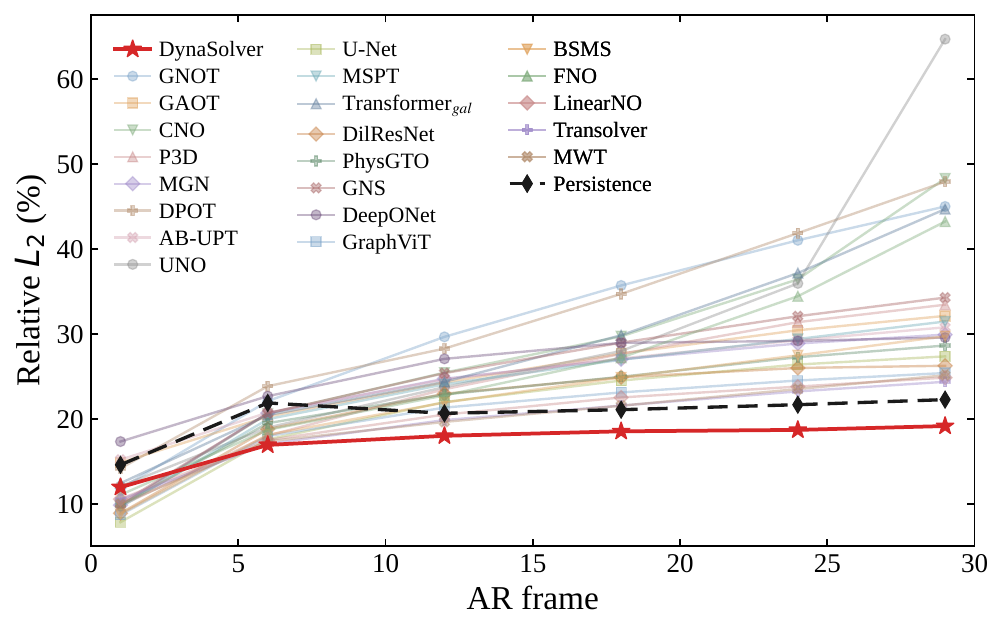}
        \caption{Overall rollout error.}
        \label{fig:rollout_all}
    \end{subfigure}
    \hfill
    \begin{subfigure}{0.32\textwidth}
        \centering
        \includegraphics[width=\linewidth]{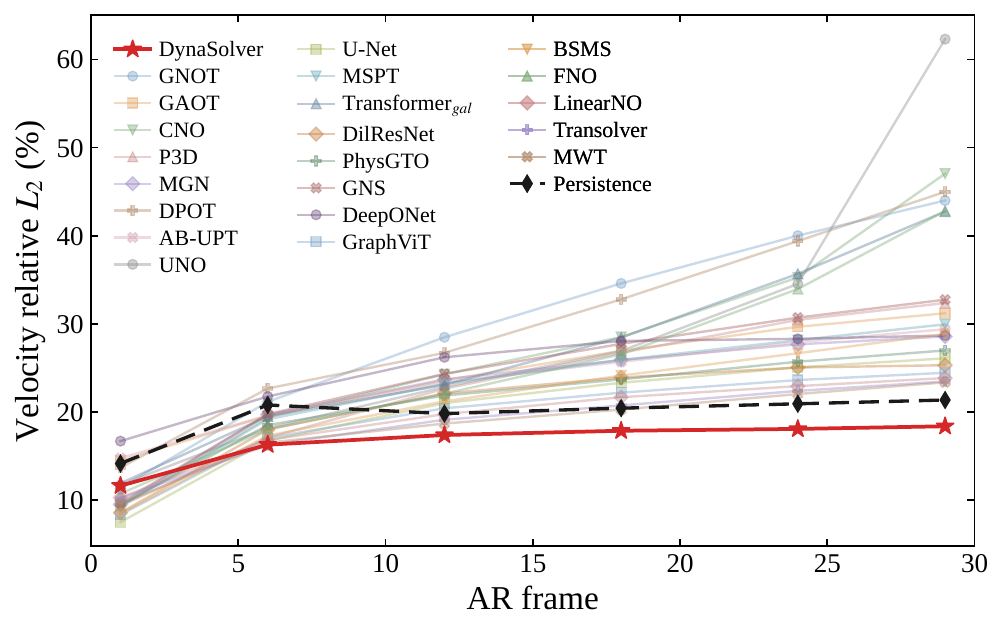}
        \caption{Velocity rollout error.}
        \label{fig:rollout_velocity}
    \end{subfigure}
    \hfill
    \begin{subfigure}{0.32\textwidth}
        \centering
        \includegraphics[width=\linewidth]{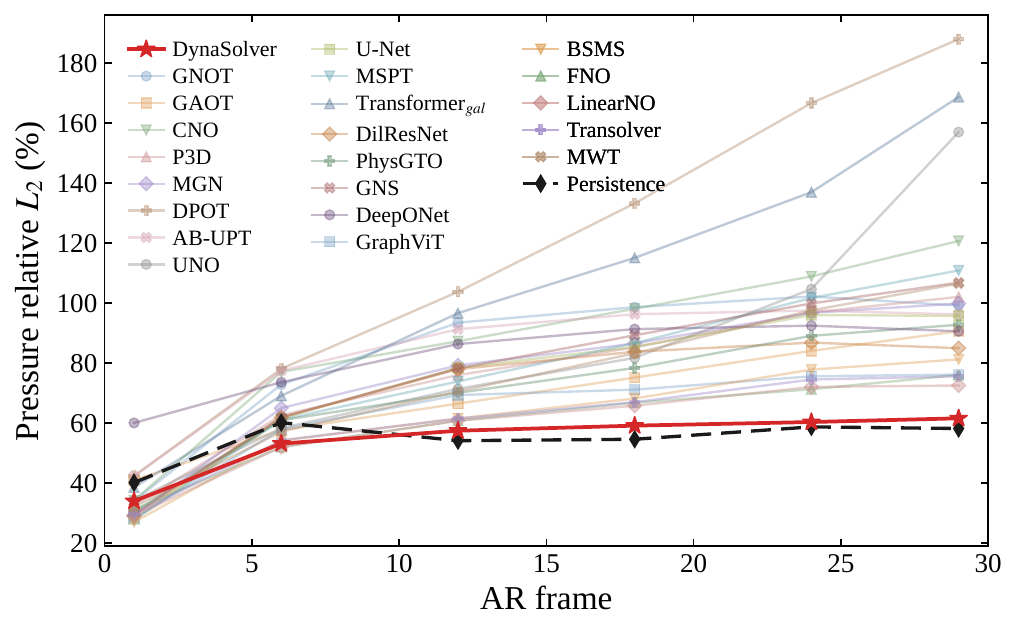}
        \caption{Pressure rollout error.}
        \label{fig:rollout_pressure}
    \end{subfigure}
    \caption{\textbf{Rollout error evolution on RoboCFD-4D Base.}
    Relative $L_2$ errors are reported across the full autoregressive rollout
    for the overall flow field, velocity, and pressure.}
    \label{fig:rollout_plot}
\end{figure*}

\begin{figure}[t]
\includegraphics[width=0.98\linewidth]{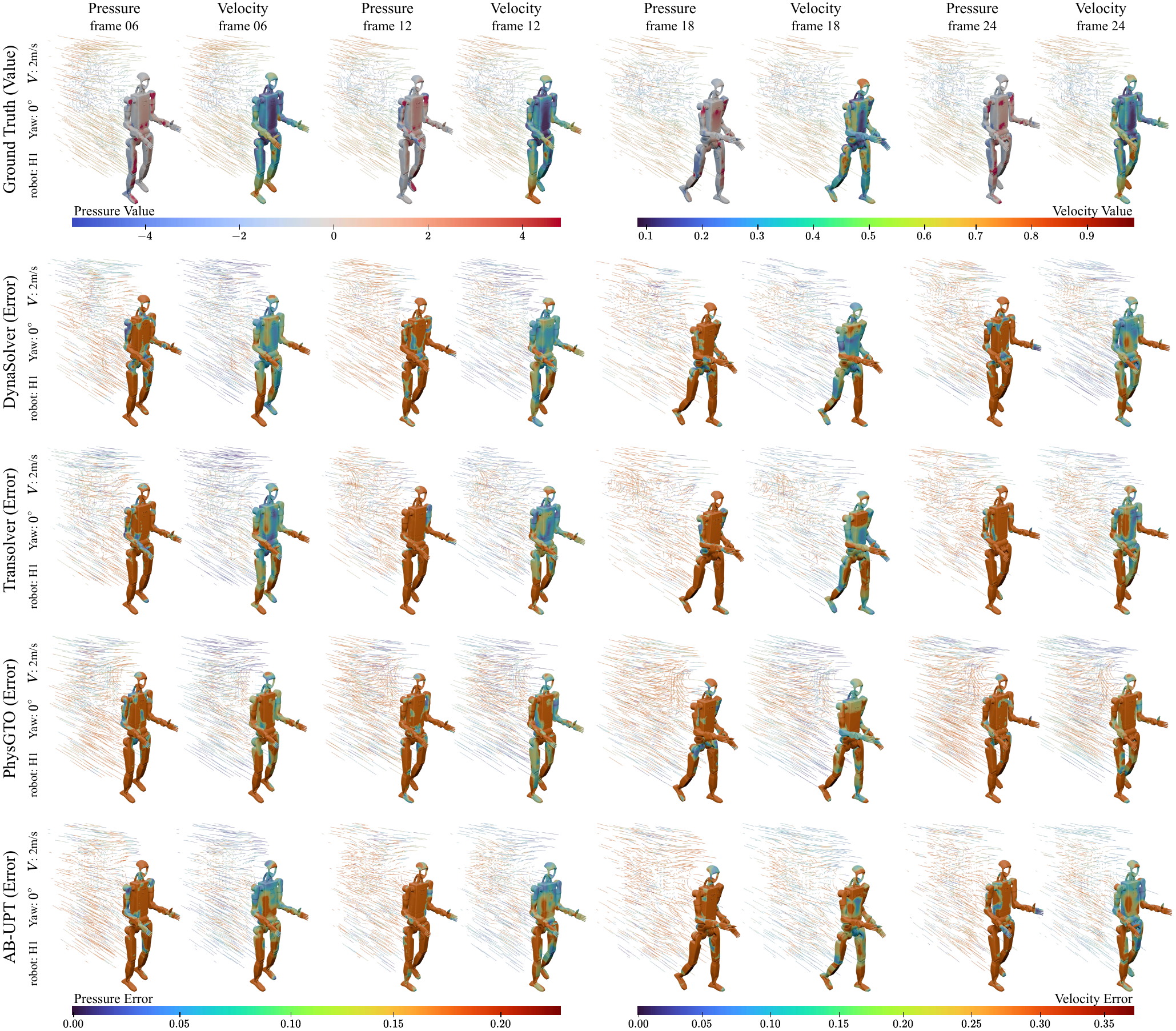}
\centering
\caption{
\textbf{Qualitative comparison across autoregressive rollout.}
Ground-truth fields and corresponding error maps are shown at selected rollout steps for an unseen robot morphology. DynaSolver shows lower long-horizon errors than the baselines, particularly for velocity.
}
\label{fig:experiment}
\end{figure}

\subsection{RoboCFD-4D Dataset}
\label{sec:dataset}

We introduce \textbf{RoboCFD-4D}, a dynamic-boundary 4D CFD benchmark for articulated robotic motion. 
Each trajectory contains $30$ temporally ordered CFD states ($T_{\mathrm{roll}}=29$), each represented on an $18\times18\times36$ model grid ($N=11{,}664$) with aligned volumetric and surface-associated features, and spans freestream velocities in $\{2,3,4,6,7,8\}\,\mathrm{m/s}$ under multiple yaw angles.
\textbf{RoboCFD-4D Base}, used for the main experiments, contains 216 training trajectories from 18 robot morphologies and 136 development/evaluation trajectories covering unseen wind conditions and four unseen morphologies.
\textbf{RoboCFD-4D Mini} is a strict subset used for controlled ablations, with 96 training trajectories from eight morphologies and 48 development/evaluation trajectories including one unseen morphology. 
Details of the simulation and dataset construction pipeline are provided in the appendix (Sec.~\ref{sec:data_generation}).

\subsection{Main Results}
\label{sec:main_results}
We compare DynaSolver with representative learning-based CFD surrogates on RoboCFD-4D Base. A non-learned persistence baseline is also included, which keeps the initial flow state unchanged throughout rollout, i.e., $\widehat{Y}_t=Y_0$, providing a no-evolution reference against which the value of modeling dynamic boundary and flow evolution can be assessed.

\begin{wrapfigure}{r}{0.38\textwidth}
    \centering
    \includegraphics[width=\linewidth]
    {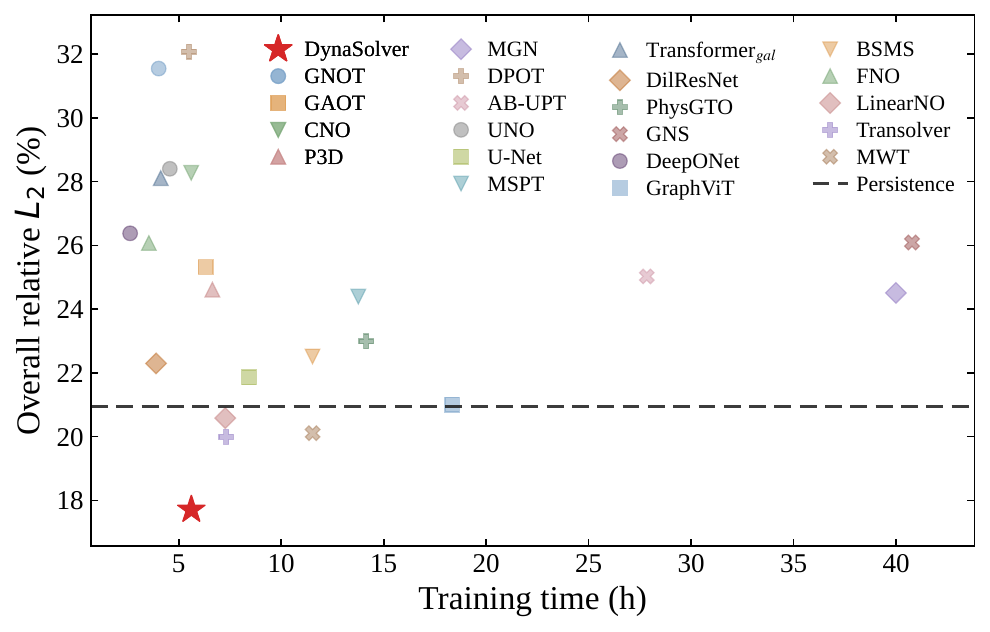}
    \caption{\textbf{Training cost versus overall rollout error.}
    Overall rollout relative $L_2$ error is plotted against training time; DynaSolver achieves the lowest rollout error while maintaining low training cost.}
    \label{fig:train_efficiency}
\end{wrapfigure}
As shown in Table~\ref{tab:rollout_comparison}, although DynaSolver does not attain the lowest one-step error, its advantage becomes more pronounced during autoregressive rollout. From step $12$ onward, it consistently achieves the lowest velocity and pressure errors among the evaluated learned methods. At the final step $t=29$, DynaSolver reaches $18.36\%$ velocity error and $61.61\%$ pressure error, compared with the best competing learned-method results of $23.35\%$ and $72.55\%$, respectively, corresponding to relative reductions of $21.4\%$ and $15.1\%$. The rollout curves in Fig.~\ref{fig:rollout_plot} and qualitative comparisons in Fig.~\ref{fig:experiment} further illustrate this advantage, particularly for velocity prediction at later rollout steps.

The advantage is also reflected in the full-rollout averages. DynaSolver achieves $17.07\%$ velocity error, $56.35\%$ pressure error, and $17.71\%$ overall error, yielding an $11.4\%$ relative reduction over the best competing learned method ($19.99\%$).
Compared with the non-learned Persistence baseline, DynaSolver achieves lower average velocity error (17.07\% vs. 20.14\%) and overall error (17.71\% vs. 20.95\%), while Persistence yields a slightly lower average pressure error (56.09\% vs. 56.35\%). This highlights that the improvements are not uniform across physical variables, despite DynaSolver's stronger long-horizon pressure prediction relative to the evaluated learned methods.
Importantly, DynaSolver maintains low training cost, achieving the lowest overall rollout error while remaining among the most training-efficient learned methods (Fig.~\ref{fig:train_efficiency}). Together, these results demonstrate improved autoregressive rollout accuracy with efficient training.

\subsection{Ablation Studies}
\label{sec:ablation}

\begin{table*}[t]
\centering
\setlength{\tabcolsep}{6pt}
\renewcommand{\arraystretch}{1.15}

\caption{\textbf{Progressive design ablation on RoboCFD-4D Mini.}
We report the full-rollout relative $L_2$ errors (\%) together with training cost, inference latency, and model size. The \colorbox[HTML]{F3F7FC}{blue} row is ours.}
\label{tab:model_ablation}

\resizebox{\textwidth}{!}{%
\begin{tabular}{l|ccc|cc|c}
\toprule[1.4pt]

\textbf{Variants}
& \textbf{Average Velocity $L_2$} $\downarrow$
& \textbf{Average Pressure $L_2$} $\downarrow$
& \textbf{Overall $L_2$} $\downarrow$
& \textbf{Train Time (h)} $\downarrow$
& \textbf{Inference Time (ms)} $\downarrow$
& \textbf{Params} \\

\midrule

Transolver
& 20.12
& 71.39
& 21.06
& \textbf{1.52}
& \textbf{8.09}
& 3.87M \\

+ Boundary Evolution
& 19.83
& 66.45
& 20.69
& \underline{1.53}
& \underline{8.14}
& 3.87M \\

+ Causal Temporal Attention
& 17.53
& 45.91
& 17.98
& 6.68
& 14.79
& 5.98M \\

+ Surface-Volume Dual Branch
& 16.24
& \textbf{42.97}
& 16.68
& 11.47
& 27.34
& 8.09M \\

+ Amortized Token-Space Operator
& 18.39
& 53.13
& 18.98
& 3.09
& 9.99
& 8.09M \\

\cellcolor[HTML]{F3F7FC}+ Pointwise MLP Decoder
& \cellcolor[HTML]{F3F7FC}\textbf{15.65}
& \cellcolor[HTML]{F3F7FC}\underline{44.30}
& \cellcolor[HTML]{F3F7FC}\textbf{16.14}
& \cellcolor[HTML]{F3F7FC}2.31
& \cellcolor[HTML]{F3F7FC}10.85
& \cellcolor[HTML]{F3F7FC}8.62M \\

- Boundary Evolution
& \underline{15.96}
& 48.88
& \underline{16.54}
& 2.31
& 10.78
& 8.62M \\

\bottomrule[1.4pt]
\end{tabular}%
}
\end{table*}

\begin{table*}[!htbp]
\centering
\setlength{\tabcolsep}{4pt}
\renewcommand{\arraystretch}{1.15}

\caption{\textbf{Training-strategy ablation on RoboCFD-4D Mini.}
Left: full-rollout relative $L_2$ errors (\%) and training cost under the same DynaSolver architecture, TF represents Teacher Forcing. Right: training convergence over epochs. The \colorbox[HTML]{F3F7FC}{blue} row is ours.}
\label{tab:training_stragety_abl}

\begin{minipage}[t]{0.71 \textwidth}
\centering

\adjustbox{valign=t}{%
\resizebox{\linewidth}{!}{%
\begin{tabular}{l|ccc|c}
\toprule[1.4pt]

\textbf{Training Strategy}
& \textbf{Avg. Vel. $L_2$} $\downarrow$
& \textbf{Avg. Pre. $L_2$} $\downarrow$
& \textbf{Overall $L_2$} $\downarrow$
& \textbf{Training Time (h)} $\downarrow$ \\

\midrule

\textcolor{gray}{Unroll}
& \textcolor{gray}{16.00}
& \textcolor{gray}{43.83}
& \textcolor{gray}{16.47}
& \textcolor{gray}{16.58} \\

\textcolor{gray}{Pushforward}
& \textcolor{gray}{16.04}
& \textcolor{gray}{44.78}
& \textcolor{gray}{16.53}
& \textcolor{gray}{16.48} \\

Parallel TF
& 17.15
& 65.08
& 18.18
& \textbf{1.79} \\

Parallel TF + Noise
& 16.45
& 52.87
& 17.14
& \underline{1.84} \\

Pure Replay
& \underline{16.17}
& \underline{48.21}
& \underline{16.72}
& {2.82} \\

\cellcolor[HTML]{F3F7FC}Replay-Augmented Parallel
& \cellcolor[HTML]{F3F7FC}\textbf{15.65}
& \cellcolor[HTML]{F3F7FC}\textbf{44.30}
& \cellcolor[HTML]{F3F7FC}\textbf{16.14}
& \cellcolor[HTML]{F3F7FC}2.31 \\

\bottomrule[1.4pt]
\end{tabular}%
}%
}

\end{minipage}
\hfill
\begin{minipage}[t]{0.27\textwidth}
\centering

\adjustbox{valign=t}{%
    \includegraphics[width=\linewidth]
    {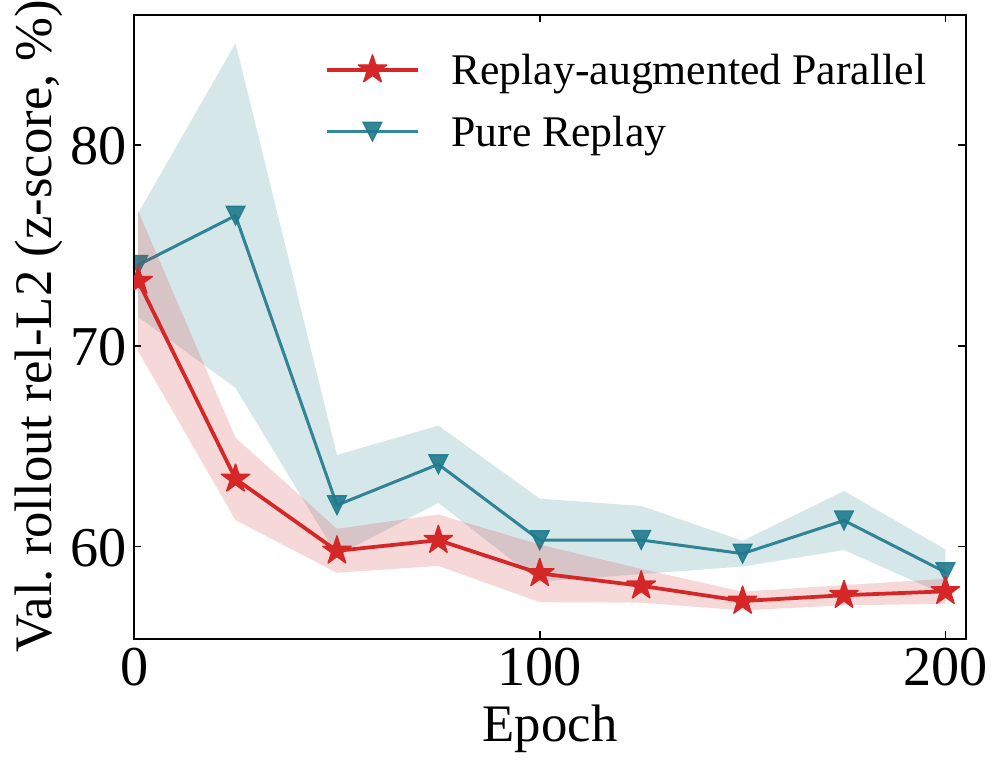}%
}

\end{minipage}
\end{table*}

\textbf{Progressive model design.} Table~\ref{tab:model_ablation} progressively builds DynaSolver from the spatial Transolver backbone. Boundary-evolution conditioning first improves sensitivity to moving-boundary dynamics, with the clearest gain in pressure prediction. Introducing causal temporal modeling then yields the largest improvement, reducing the overall rollout error from $20.69\%$ to $17.98\%$, while the dual surface--volume branch further lowers it to $16.68\%$. We next move the repeated computation into the compact token space, which sharply reduces the training cost from $11.47$ h to $3.09$ h but sacrifices accuracy when used alone. 
The pointwise MLP decoder restores point-level expressiveness, reducing the overall error from 18.98\% to 16.14\% while preserving the computational efficiency of the token-space operator; the complete DynaSolver achieves the best overall error of $16.14\%$ with only $2.31$ h of training.

Notably, removing boundary-evolution conditioning increases the overall error only to 16.54\%, still outperforming Transolver, the strongest learned baseline, under matched input conditions. This demonstrates that DynaSolver's improvements are not solely attributable to additional boundary information.

\begin{table*}[t]
\centering
\setlength{\tabcolsep}{3.5pt}
\renewcommand{\arraystretch}{1.12}

\caption{
\textbf{Sensitivity Analysis on RoboCFD-4D Mini.}
We vary the temporal window length $W$ while fixing the slice count at $S=32$ (left), and vary $S$ while fixing $W=4$ (right).
The \colorbox[HTML]{F3F7FC}{blue} row is ours.
}
\label{tab:window_slice_select}

\noindent
\parbox[c]{0.48\textwidth}{%
\vspace{0pt}
\centering
\resizebox{\linewidth}{!}{%
\begin{tabular}{c|ccc|c}
\toprule[1.2pt]
\textbf{$W$}
& \textbf{Avg.\ Vel.\ $L_2$} $\downarrow$
& \textbf{Avg.\ Pre.\ $L_2$} $\downarrow$
& \textbf{Overall $L_2$} $\downarrow$
& \textbf{Train (h)} $\downarrow$ \\
\midrule

1 &
18.27 &
66.17 &
19.22 &
\textbf{0.98} \\

2 &
16.10 &
48.52 &
16.67 &
\underline{1.06} \\

3 &
16.04 &
48.11 &
16.60 &
1.53 \\

\rowcolor[HTML]{F3F7FC}4 &
\textbf{15.65} &
\textbf{44.30} &
\textbf{16.14} &
2.31 \\

5 &
15.79 &
\underline{45.48} &
16.29 &
3.07 \\

6 &
\underline{15.75} &
45.89 &
\underline{16.28} &
3.21 \\

\bottomrule[1.2pt]
\end{tabular}%
}
}%
\hfill%
\parbox[c]{0.495\textwidth}{%
\vspace{0pt}
\centering
\resizebox{\linewidth}{!}{%
\begin{tabular}{c|ccc|c}
\toprule[1.2pt]
\textbf{$S$}
& \textbf{Avg.\ Vel.\ $L_2$} $\downarrow$
& \textbf{Avg.\ Pre.\ $L_2$} $\downarrow$
& \textbf{Overall $L_2$} $\downarrow$
& \textbf{Train (h)} $\downarrow$ \\
\midrule

4 &
15.68 &
45.93 &
\underline{16.20} &
\textbf{1.29} \\

8 &
\underline{15.67} &
47.60 &
16.23 &
\underline{1.61} \\

16 &
15.81 &
\textbf{42.54} &
16.24 &
1.95 \\

\rowcolor[HTML]{F3F7FC}32 &
\textbf{15.65} &
\underline{44.30} &
\textbf{16.14} &
2.31 \\

64 &
16.10 &
45.15 &
16.58 &
2.96 \\

128 &
16.05 &
44.53 &
16.53 &
3.65 \\

\bottomrule[1.2pt]
\end{tabular}%
}
}
\end{table*}

\textbf{Training strategy ablation.} Table~\ref{tab:training_stragety_abl} fixes the complete DynaSolver architecture and isolates the effect of training strategy. Dense autoregressive schemes such as Unroll and Pushforward achieve strong rollout accuracy, but require costly sequential rollout during training. Parallel TF instead exploits block-causal attention to optimize all transitions in a clip simultaneously and is substantially faster, yet its $18.18\%$ overall error reflects the mismatch between ground-truth-conditioned training and autoregressive rollout. Injecting Gaussian noise~\citep{Pfaff2021MeshGraphNet} partially alleviates this mismatch, reducing the error to $17.14\%$, while Pure Replay further improves it to $16.72\%$ by exposing the model to autoregressively generated states.
Replay-Augmented Parallel Training combines replay histories with clean ground-truth histories while retaining parallel optimization. It achieves comparable, slightly better rollout accuracy than Unroll (16.47\%) and Pushforward (16.53\%), reaching 16.14\% overall error with approximately $7.1\times$ lower training time (2.31 h). The convergence curves further show lower validation rollout error than Pure Replay across most training epochs.

\textbf{Sensitivity Analysis of Temporal Window and Slice Count.}
Table~\ref{tab:window_slice_select} examines the effects of temporal window length $W$ and slice count $S$ on rollout accuracy and training cost. Increasing $W$ from 1 to 4 reduces the overall error from $19.22\%$ to $16.14\%$, highlighting the benefit of incorporating temporal context. Further increasing $W$ yields no additional accuracy gains while increasing training time. For slice count, $S=32$ achieves the lowest overall error of $16.14\%$, while larger values increase training cost and degrade accuracy. We therefore adopt $W=4$ and $S=32$ as the default configuration.

%% file: sections/04_conclusion.tex
\section{Conclusion}
\label{sec:conclusion}

We presented DynaSolver for 4D CFD prediction with continuously evolving robot boundaries. DynaSolver models causal spatio-temporal flow evolution under dynamic boundaries through an efficient physical-token temporal operator, while Replay-Augmented Parallel Training exposes the model to autoregressive histories while preserving efficient parallel optimization. Experiments on RoboCFD-4D show that DynaSolver achieves strong prediction accuracy and training efficiency, with its advantage becoming increasingly pronounced as rollout proceeds. These results demonstrate reduced autoregressive error accumulation and support causal spatio-temporal modeling as an effective approach to neural CFD prediction under dynamic geometry.

%% file: sections/05_appendix.tex
\appendix
\label{sec:appendix}

\section{Experimental Setup}
\label{sec:experimental_setup}
\textbf{Input representation.}
For each transition from $t-1$ to $t$, DynaSolver receives the spatial condition $c_t\in\mathbb{R}^{N\times11}$, the boundary-evolution feature $b_{t-1\rightarrow t}\in\mathbb{R}^{N\times8}$, and the previous CFD state $Y_{t-1}\in\mathbb{R}^{N\times4}$. Table~\ref{tab:input_features} summarizes the channel definitions. 
All features are defined pointwise on the model grid, with surface-associated attributes provided at the corresponding projected grid nodes.
The signed-distance convention is $\phi>0$ in the fluid domain and $\phi<0$ inside the solid, and surface normals point outward from the robot. 
All baselines receive the spatial condition $c_t$ and previous CFD state $Y_{t-1}$. DynaSolver additionally incorporates the proposed boundary-evolution representation $b_{t-1\rightarrow t}$; its contribution is isolated in the progressive model ablation in Table~\ref{tab:model_ablation}.

\begin{table}[!htbp]
\centering
\caption{\textbf{Pointwise input features used by DynaSolver.}
$\mathrm{S}$ and $\mathrm{V}$ denote surface-associated and volume grid nodes, respectively.}
\label{tab:input_features}
\small
\setlength{\tabcolsep}{4pt}
\begin{tabular}{llccl}
\toprule
\textbf{Input} & \textbf{Feature} & \textbf{Dim.} & \textbf{Samples} & \textbf{Representation} \\
\midrule
$c_t$
& normalized position $\widetilde{\mathbf{x}}_t$
& 3 & S/V & dimensionless \\

& signed distance $\phi_t$
& 1 & S/V & m \\

& nearest outward normal $\mathbf{n}_t$
& 3 & S/V & unit vector \\

& inflow direction $\mathbf{d}_{\infty}$
& 3 & S/V & unit vector \\

& normalized inflow speed $\widetilde U_{\infty}$
& 1 & S/V & dimensionless \\
\midrule
$b_{t-1\rightarrow t}$
& previous signed distance $\phi_{t-1}$
& 1 & S/V & m \\

& signed-distance change $\phi_t-\phi_{t-1}$
& 1 & S/V & m \\

& wall velocity $\mathbf{v}^{\mathrm{w}}_{t-1\rightarrow t}$
& 3 & S/V & m/s \\

& fluid indicators $(\chi_{t-1},\chi_t)$
& 2 & S/V & binary \\

& normalized time interval $\widetilde{\Delta t}$
& 1 & S/V & dimensionless \\
\bottomrule
\end{tabular}
\end{table}

\textbf{Model and training configuration.} Our default DynaSolver uses a latent width of $D=256$, $L=8$ operator blocks with eight attention heads, $G=32$ physics tokens per surface/volume branch, and a causal window of $W=4$. We sample non-overlapping temporal clips with stride $4$ and train all models from scratch for 200 epochs with batch size $4$, learning rate $10^{-4}$, and weight decay $10^{-5}$. For Replay-Augmented Parallel Training, the replay probability $\rho$ is gradually increased from $0$ to $0.5$ over the first 20 epochs and kept fixed thereafter. All reported training and inference times are measured on a single NVIDIA A800 GPU. Base is used for the main comparison, while Mini is used for controlled ablations under the same model and optimization configuration.

\textbf{Evaluation protocol and metrics.}
Following the formulation in Sec.~\ref{sec:problem}, we evaluate all models under closed-loop autoregressive rollout over the full horizon $T_{\mathrm{roll}}$, starting from the given initial state $Y_0$ and recursively feeding each prediction back into the subsequent causal context.

The CFD state is $Y_t=[u,v,w,p]_t$, where $\mathbf{u}=(u,v,w)$ denotes the velocity vector and $p$ denotes gauge-referenced kinematic pressure. For each trajectory, let $U_\infty$ denote the magnitude of the prescribed freestream velocity. We first nondimensionalize the flow variables as
\begin{equation}
    \mathbf{u}^{*}=\frac{\mathbf{u}}{U_\infty},
    \qquad
    p^{*}=\frac{p}{U_\infty^{2}},
    \label{eq:flow_nondim}
\end{equation}
where $(\cdot)^{*}$ denotes the corresponding nondimensional quantity. Each channel is then standardized using its mean and standard deviation computed exclusively from the training split. Denoting this standardization by $\mathcal{N}(\cdot)$, the model is trained and autoregressively rolled out in the standardized space,
\begin{equation}
    Z_t=\mathcal{N}(Y_t^{*}).
\end{equation}
For evaluation, predicted states are first transformed back to the nondimensional space as $\widehat{Y}_t^{*}=\mathcal{N}^{-1}(\widehat{Z}_t)$. For a channel subset $\mathcal{S}$, the masked relative $L_2$ error is
\begin{equation}
    E_{\mathcal{S}}(t)
    =
    \sqrt{
    \frac{
        \sum_{n=1}^{N}
        m_{t,n}
        \sum_{c\in\mathcal{S}}
        \left(
            \widehat{Y}^{*}_{t,n,c}
            -
            Y^{*}_{t,n,c}
        \right)^2
    }{
        \sum_{n=1}^{N}
        m_{t,n}
        \sum_{c\in\mathcal{S}}
        \left(
            Y^{*}_{t,n,c}
        \right)^2
    }}.
    \label{eq:relative_l2_suppl}
\end{equation}
where $m_{t,n}$ excludes invalid samples from evaluation, including volume samples occupied by the moving solid according to $M_t^{\mathrm v}$. We report velocity over $\mathcal{S}_{v}=\{u,v,w\}$, pressure over $\mathcal{S}_{p}=\{p\}$, and the overall field over $\mathcal{S}_{\mathrm{all}}=\{u,v,w,p\}$, with all errors expressed as percentages. 
Computing the metric in nondimensional space removes dependence on training-set standardization and avoids scale bias across variables.

Horizon-wise results report $E_{\mathcal{S}}(t)$ at selected rollout steps. The full-rollout score first averages the per-step errors over $T_{\mathrm{roll}}$ within each trajectory and then across trajectories. 
RoboCFD-4D uses a train/test protocol without a separate validation split.
No checkpoint selection is performed on the test set; all models are trained for
the fixed 200-epoch schedule described above, and the final-epoch checkpoint is
used for evaluation. The same training and evaluation protocol is applied to all
compared methods.

\section{RoboCFD-4D}
\label{sec:data_generation}
We generate RoboCFD-4D by simulating flow around articulated robots undergoing prescribed motion. For each robot, a fixed 1 s motion sequence is evaluated under 16 prescribed wind conditions, with 200 raw CFD output frames stored for each completed condition. After removing the initial moving-boundary start-up transient, we construct temporally subsampled learning sequences from these outputs. Each retained frame contains both Eulerian flow fields and observations associated with a fixed set of points on the moving robot. Throughout simulation and export, versioned manifests and stable link indices preserve a consistent correspondence among geometry, motion, and flow observations throughout the dataset.

\paragraph{Public assets and link preparation.}
We obtain robot descriptions and meshes from public repositories,
deduplicate overlapping descriptions, and convert URDF and MJCF assets
to a common URDF-based articulation representation. This representation
records the kinematic tree, joint axes and limits, and the geometry of
each link in its local coordinate frame. After resolving mesh
dependencies, we validate each link mesh independently. Meshes that fail
geometry checks are repaired at the link level, with CGAL Alpha Wrap
used where needed. Repairing geometry in the original link frame
preserves its attachment to the kinematic tree and its correspondence
with the subsequent link-local SDF representation. Before simulation,
the resulting CFD inputs are checked for watertightness, winding
consistency, and geometric validity.

\paragraph{Kinematic motion synthesis.}
We use two forms of prescribed kinematic motion. Most articulated joints
are handled by a generic morphology-aware synthesis pipeline, whereas
robot-specific profiles are used for mechanisms that require dedicated
treatment, including continuous wheel rotations. For joints handled by
the generic pipeline, motion follows the harmonic template
\begin{equation}
q_j(t)
=
\bar{q}_j
+
a_j \sin(2\pi f t + \phi_j)
+
b_j \cos(2\pi f t + \phi_j),
\label{eq:joint_motion}
\end{equation}
where $\bar{q}_j$ denotes the neutral pose, $f$ is the motion frequency,
and $a_j$, $b_j$, and $\phi_j$ control the excursion and relative timing
of joint $j$. Generic-motion amplitudes are selected to satisfy joint
position and velocity limits, followed by kinematic consistency checks.
The root undergoes no net horizontal locomotion, although bounded
lateral, vertical, and angular oscillations may be prescribed. The
resulting trajectories provide smooth, kinematically valid articulated
boundary motion rather than dynamically simulated locomotion.

Each source motion is sampled at $1\,\mathrm{ms}$ intervals over
$[0,1]\,\mathrm{s}$, yielding $1001$ source samples. Forward kinematics
provides the transform $\mathbf{T}_{\ell}(t)\in\mathrm{SE}(3)$ of each
link $\ell$. Raw CFD outputs are stored every fifth source interval, at
$t=0.005,0.010,\ldots,1.000\,\mathrm{s}$, yielding $200$ frames per
completed condition. Generic motion packages undergo joint-limit,
temporal-closure, rotation, forward-kinematics, foot-clearance, and
discrete bounding-box checks where applicable. Dedicated prescribed
motion packages are audited separately for temporal and articulation
compatibility with the CFD pipeline.

\paragraph{Dynamic geometry reconstruction.}
To provide the moving geometry to the flow solver, we construct a
truncated signed distance field (SDF) $\phi_{\ell}$ for each repaired
link in its local coordinate system. The transformed link-local fields
are combined into the composite implicit representation
\begin{equation}
\phi(\mathbf{x},t)
=
\min_{\ell}
\phi_{\ell}
\left(
\mathbf{T}_{\ell}^{-1}(t)\mathbf{x}
\right),
\label{eq:articulated_sdf}
\end{equation}
where $\mathbf{x}$ is a world-space point, homogeneous coordinates are
implicit, and negative values indicate solid interiors. The negative
region of the composite field represents the union of the transformed
link solids. The pointwise minimum is not required to remain an exact
signed-distance function everywhere, particularly near link
interfaces.

Small overlaps may occur between tightly fitted neighboring parts due
to finite precision and discretization in the source meshes. Under the
pointwise minimum, such overlaps are treated as part of a single solid
union rather than as separate internal boundaries. They therefore
reflect mesh precision at close-fitting interfaces rather than intended
interpenetration between distinct robot components.

Each link-local SDF is precomputed once and reused throughout the
sequence. At runtime, only the rigid link transforms are updated, and
the composite field is evaluated from the transformed link-local fields
using Eq.~\ref{eq:articulated_sdf}. This avoids reconstructing a
volumetric geometry representation at every frame while preserving
link-level geometry and articulation.

Each link-local SDF is validated using surface reconstruction error, inside--outside sign accuracy, and voxelized-volume agreement. Fields that fail these gates are rebuilt and revalidated before use. At representative poses, we additionally check link transforms and per-link mesh--SDF consistency. Separate domain-fit checks verify that the swept moving geometry remains inside the prescribed computational domain throughout the simulated motion.

\begin{figure*}[t]
\centering
\includegraphics[width=\textwidth]{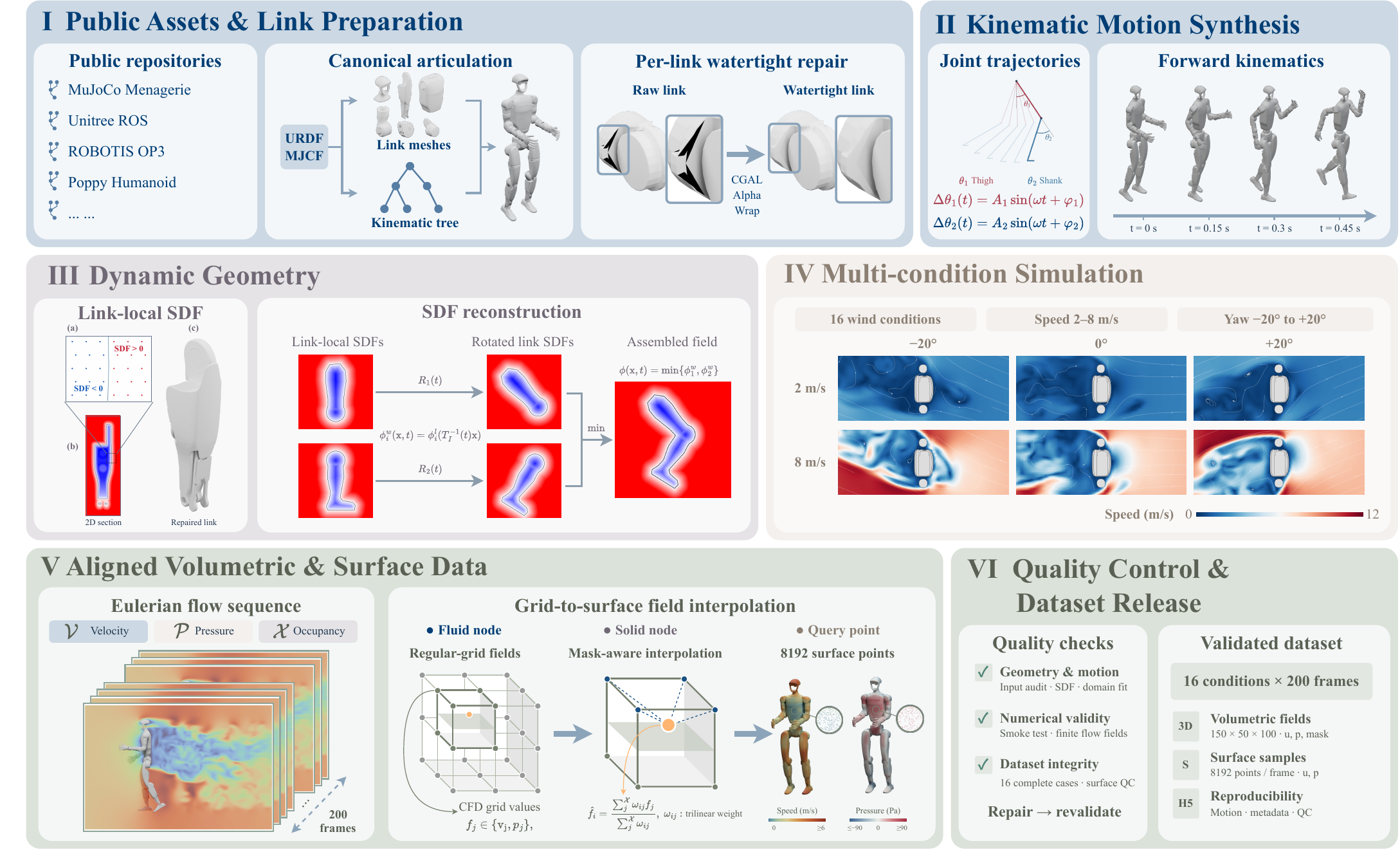}
\caption{Overview of the RoboCFD-4D data generation pipeline. Public robot assets are standardized and repaired at the link level, followed by prescribed motion synthesis and reusable link-local SDF construction. The transformed link-local fields define the prescribed moving immersed boundary for multi-condition CFD simulations. Aligned volumetric fields and tracked near-surface observations are then extracted and subjected to geometric, numerical, and dataset-level validation before inclusion in the final published dataset.
}
\label{fig:robot_cfd_pipeline}
\end{figure*}

\paragraph{Multi-condition flow simulation.}
We use WaterLily~\citep{Weymouth_2025} to solve the incompressible Navier--Stokes equations, with the articulated robot geometry defining a prescribed moving immersed boundary throughout the simulation:

\begin{align}
\nabla\cdot\mathbf{u} &= 0, \\
\frac{\partial\mathbf{u}}{\partial t}
+
(\mathbf{u}\cdot\nabla)\mathbf{u}
&=
-\nabla p
+
\nu\nabla^2\mathbf{u},
\label{eq:navier_stokes}
\end{align}
where $\mathbf{u}$ is velocity, $p$ is kinematic pressure, and $\nu$ is
kinematic viscosity. The composite articulated field and prescribed
link motion determine the moving solid boundary throughout the
simulation; no two-way fluid--structure coupling is used.

The $16$ prescribed wind conditions are listed in
Table~\ref{tab:wind_conditions}.

\begin{table}[t]
\centering
\caption{Wind conditions and repository identifiers used in RoboCFD-4D.}
\label{tab:wind_conditions}
\begin{tabular}{ccc}
\toprule
\textbf{Speed (m/s)} & \textbf{Yaw} & \textbf{Condition ID} \\
\midrule
2 & $-20^\circ$ & \texttt{u2\_yaw\_m20} \\
2 & $0^\circ$   & \texttt{u2\_yaw\_p0} \\
2 & $20^\circ$  & \texttt{u2\_yaw\_p20} \\
3 & $-10^\circ$ & \texttt{u3\_yaw\_m10} \\
3 & $10^\circ$  & \texttt{u3\_yaw\_p10} \\
4 & $-20^\circ$ & \texttt{u4\_yaw\_m20} \\
4 & $0^\circ$   & \texttt{u4\_yaw\_p0} \\
4 & $20^\circ$  & \texttt{u4\_yaw\_p20} \\
6 & $-20^\circ$ & \texttt{u6\_yaw\_m20} \\
6 & $0^\circ$   & \texttt{u6\_yaw\_p0} \\
6 & $20^\circ$  & \texttt{u6\_yaw\_p20} \\
7 & $-10^\circ$ & \texttt{u7\_yaw\_m10} \\
7 & $10^\circ$  & \texttt{u7\_yaw\_p10} \\
8 & $-20^\circ$ & \texttt{u8\_yaw\_m20} \\
8 & $0^\circ$   & \texttt{u8\_yaw\_p0} \\
8 & $20^\circ$  & \texttt{u8\_yaw\_p20} \\
\bottomrule
\end{tabular}
\end{table}

All simulations use a $150\times50\times100$ grid with isotropic
physical cell spacing. Grid dimensions are fixed across robots, whereas
the physical cell size $\Delta x$ is selected separately for each robot
to accommodate its swept geometry over the prescribed motion and wind
orientations. Thus, simulations share the same grid dimensions but not
necessarily the same physical spatial resolution or domain size. Across the dataset, the swept vertical extent spans
approximately $38$--$77$ grid cells.

Before each full run, a short preliminary simulation checks geometry--solver integration and basic numerical consistency. During the dynamic segment, each wind condition uses a fixed time step selected from the grid spacing and motion-speed estimates. CFL ratios are monitored during execution, and normalized velocity divergence and flux imbalance are evaluated as part of the numerical quality-control procedure described below.

\paragraph{Aligned volumetric and near-surface observations.}
Each completed simulation produces $200$ raw CFD output frames indexed
from $0$ to $199$. Raw frame $k$ corresponds to
$t=(k+1)\times0.005\,\mathrm{s}$, so frame $0$ is stored at
$t=0.005\,\mathrm{s}$. The earliest outputs immediately follow the
transition from the stationary-body flow initialization to the
prescribed moving-body simulation and therefore contain a pronounced
start-up transient.

To reduce the influence of this transition regime on the learning
benchmark, we discard raw frames $0$--$19$ and uniformly subsample the
remaining $180$ frames with a stride of $6$. The retained zero-based
raw-frame indices are
\[
20,26,32,\ldots,194,
\]
corresponding to
\[
t=0.105,0.135,0.165,\ldots,0.975\,\mathrm{s}.
\]
This yields $30$ temporally ordered frames per sequence, or $29$
one-step transitions for autoregressive learning. The same temporal
indices are applied to all volumetric fields, link transforms, tracked
surface points, and near-surface observations, thereby ensuring
consistent temporal alignment across all exported data modalities
throughout each learning sequence.

For each raw CFD output, we store the cell-centered velocity vector,
speed magnitude, gauge-normalized kinematic pressure, and fluid mask on
the Eulerian grid. We denote the fluid mask by $\chi$, with $\chi=1$ in
the fluid and $\chi=0$ in the solid. Velocity and pressure values in
solid cells are stored as NaN for downstream processing.

Pressure is reported under a fixed gauge convention. At each output
frame, we subtract from the full pressure field the mean pressure over a
predefined set of valid fluid probes near the outlet side of the
wind-aligned simulation grid. This operation fixes only the additive
pressure reference; it is not imposed as a physical outlet-pressure
boundary condition.

The gauge-probe candidates lie on the last two $x$-planes of the
wind-aligned simulation grid. Along the $y$ and $z$ directions,
candidate grid points are selected starting from the first cell with a
stride of four cells. For the $150\times50\times100$ grid, this yields
at most
\[
2\times13\times25=650
\]
candidate probes per frame. Only candidates classified as fluid by the
current-frame fluid mask are retained. Let $\mathcal{P}_t$ denote this
valid probe set. The reported gauge-normalized kinematic pressure is
\begin{equation}
p_g(\mathbf{x},t)
=
p(\mathbf{x},t)
-
\frac{1}{|\mathcal{P}_t|}
\sum_{\mathbf{x}_j\in\mathcal{P}_t}
p(\mathbf{x}_j,t).
\label{eq:pressure_gauge}
\end{equation}
Export fails if $\mathcal{P}_t$ is empty. The residual mean of $p_g$
over $\mathcal{P}_t$ is subsequently required to be no greater than
$10^{-4}$ in the stored kinematic-pressure units. Probe locations are
fixed in the wind-aligned simulation-grid frame and therefore do not
correspond to identical world-frame locations across different yaw
conditions. We do not assume that the probe pressure equals ambient
pressure or that the far-field pressure is zero. The stored volumetric
pressure field $p_g$ has units of $\mathrm{m^2\,s^{-2}}$.

We additionally construct a deterministic bank of $8192$ tracked
surface points for each robot. Each link receives a minimum allocation,
with the remaining points distributed according to link surface area.
The local coordinates and link identities of these points are fixed
once and reused across all frames and wind conditions. Surface point
$i$ therefore follows its associated link according to
\begin{equation}
\mathbf{x}_i(t)
=
\mathbf{T}_{\ell(i)}(t)
\mathbf{x}^{\mathrm{local}}_i,
\label{eq:surface_tracking}
\end{equation}
where $\ell(i)$ denotes the link associated with point $i$. Because both
$\mathbf{x}^{\mathrm{local}}_i$ and $\ell(i)$ remain fixed, a point index
identifies the same material location throughout the motion sequence and
across wind conditions. This construction provides deterministic
material-point correspondence while retaining a fixed surface-array
layout. We also retain transformed surface normals and material-point
velocities obtained from rigid-body kinematics under the corresponding
link motion.

Flow observations are transferred from the regular grid to the tracked surface points using mask-aware trilinear interpolation. For each tracked surface point with outward normal $\mathbf{n}_i(t)$, the corresponding external flow field is queried at

\begin{equation}
\mathbf{x}^{\mathrm{query}}_i(t)
=
\mathbf{x}_i(t)
+
0.5\,\Delta x\,\mathbf{n}_i(t),
\label{eq:surface_query}
\end{equation}
where $\Delta x$ is the scalar spacing of the isotropic, cell-centered
simulation grid. The half-cell displacement is a fixed, reproducible
sampling convention that places the query on the exterior side of the
moving surface. It is not an optimized offset, and we did not perform a
dedicated offset ablation to establish $0.5\,\Delta x$ as optimal.

After mapping the query position to the cell-centered simulation grid,
a scalar field or velocity component $f$ is interpolated as
\begin{equation}
\widehat{f}_i(t)
=
\frac{
\sum_{j\in\mathcal{N}_i(t)}
m_j(t)w_{ij}(t)f_j(t)
}{
\sum_{j\in\mathcal{N}_i(t)}
m_j(t)w_{ij}(t)
},
\label{eq:masked_surface_interpolation}
\end{equation}
where $\mathcal{N}_i(t)$ contains the eight neighboring cell-centered
grid samples, $w_{ij}(t)$ are the corresponding trilinear weights, and
$m_j(t)=1$ for finite fluid samples and $0$ otherwise. Solid or
non-finite neighbors are excluded from both sums, and the remaining
weights are renormalized. A query is marked invalid and stored as NaN if
the denominator in Eq.~\ref{eq:masked_surface_interpolation} contains
no positive valid-fluid weight. For quality control, we retain both an
interpolation-validity mask and the number of valid fluid corners in the
local stencil.

These quantities are near-surface external-flow observations rather
than exact wall boundary values. Flow velocity is stored separately
from material-point velocity, and near-surface flow vectors are
expressed in world coordinates. Near-surface pressure is sampled from
the gauge-normalized kinematic pressure $p_g$ and is also provided in
pascals using
\begin{equation}
p_{\mathrm{Pa}}=\rho p_g,
\label{eq:pressure_pa}
\end{equation}
with $\rho=1.225\,\mathrm{kg\,m^{-3}}$.

\paragraph{Model-grid construction for learning.}
The raw Eulerian field contains $150\times50\times100=750{,}000$
cell-centered samples per frame. For model training and evaluation, we
do not randomly or importance-sample a subset of these cells. Instead,
for each CFD condition we construct a deterministic, cell-centered
$18\times18\times36$ grid over the physical bounds of the corresponding
simulation domain and resample the Eulerian velocity, pressure, and
fluid mask onto this grid using trilinear interpolation. This produces
exactly
\[
N=18\times18\times36=11{,}664
\]
model nodes per frame. The coarse spatial grid is fixed for all retained
frames of a given sequence, while its physical coordinates are inherited
from the corresponding CFD domain; consequently, different robots or
wind conditions are not assumed to share identical world-space node
coordinates. The same deterministic preprocessing is used for both
training and evaluation, and the current benchmark configuration uses
all $11{,}664$ nodes rather than resampling a different point subset at
each frame or evaluation pass.

The $8192$ tracked near-surface samples described above are not appended
to these $11{,}664$ nodes as additional model points. Instead, valid
tracked surface samples are projected onto cells of the same
$18\times18\times36$ model grid, where they provide the surface mask,
surface normals, material velocity, and the corresponding
surface-associated loss weights. Multiple tracked samples may map to
the same coarse-grid cell and therefore collapse onto a single model
node. Thus, $N=11{,}664$ denotes the complete structured model grid,
not a concatenation of $8192$ surface points with an independently
sampled set of volume points.

\begin{figure*}[!h]
    \centering
    \includegraphics[width=0.95\textwidth]{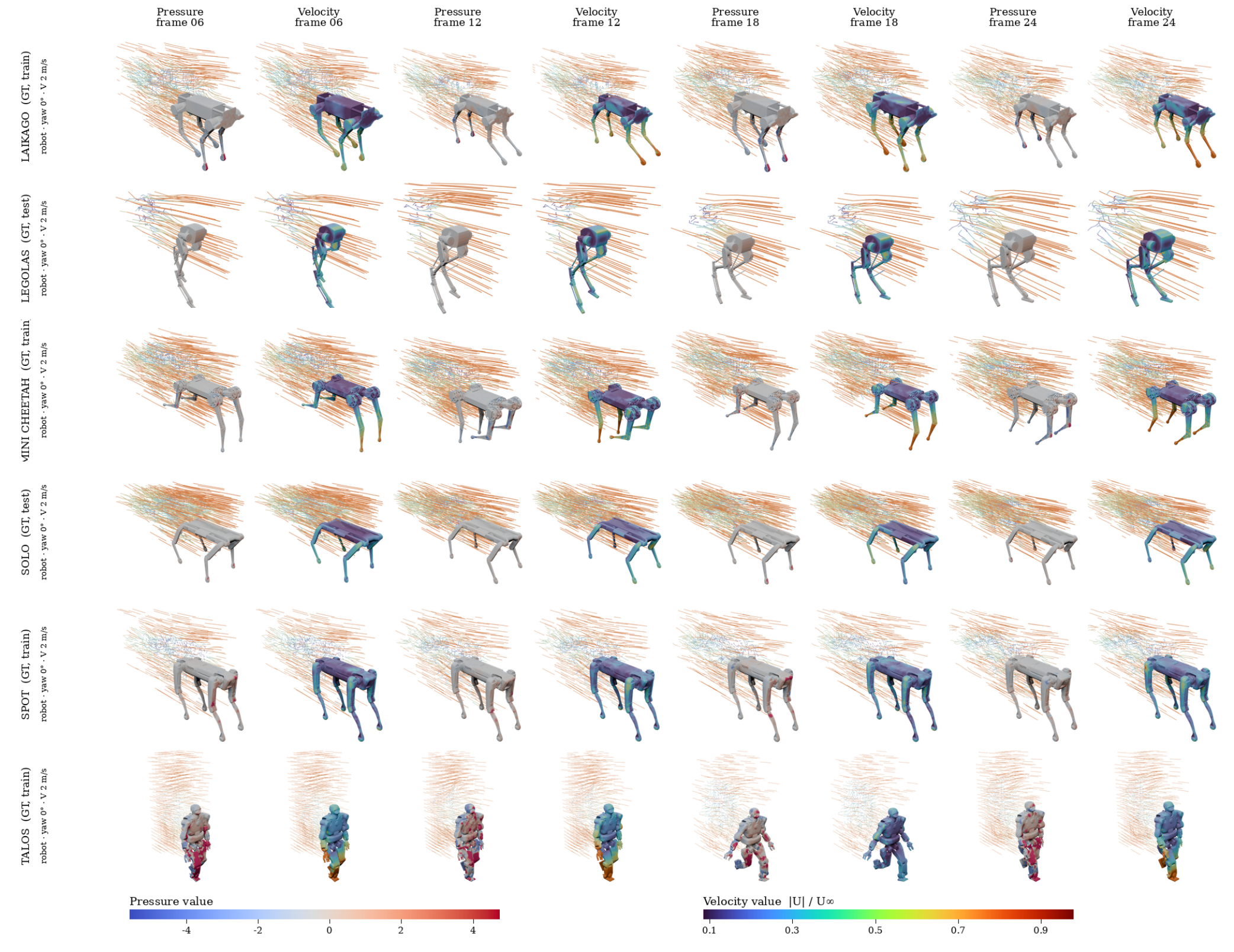}
    \caption{Ground-truth flow visualization from RoboCFD-4D. Six articulated robots are shown at rollout frames 6, 12, 18, and 24 under a 2~m/s inflow and $0^\circ$ yaw. Each pair of columns shows pressure and normalized velocity magnitude $|\mathbf{U}|/U_\infty$ around the moving robot. Row labels indicate the robot and whether its morphology is included in training or held out for evaluation.}
\label{fig:gt_01}
\end{figure*}

\begin{figure*}[!h]
    \centering
    \includegraphics[width=0.95\textwidth]{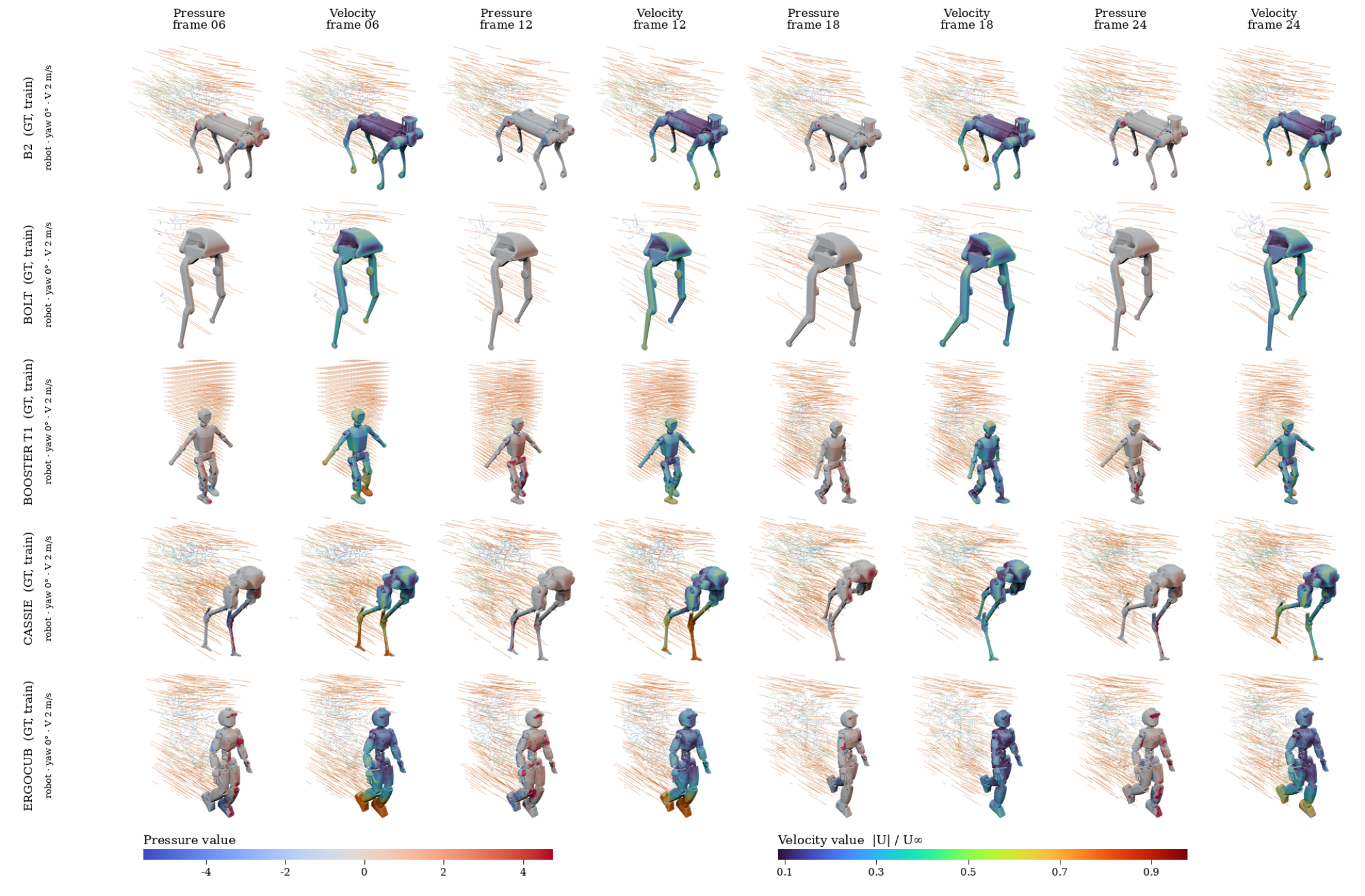}
    \caption{Ground-truth flow visualization from RoboCFD-4D. Five articulated robots are shown at rollout frames 6, 12, 18, and 24 under a 2~m/s inflow and $0^\circ$ yaw. Each pair of columns shows pressure and normalized velocity magnitude $|\mathbf{U}|/U_\infty$ around the moving robot. Row labels indicate the robot and whether its morphology is included in training or held out for evaluation.}
\label{fig:gt_02}
\end{figure*}

\paragraph{Quality control and dataset release.}
Quality control is applied at three complementary levels. Geometry and motion checks
cover mesh validity, link-local SDF consistency, articulation, and
swept-domain fit throughout the prescribed motion. Numerical checks cover temporal completeness, finite
values in fluid cells, consistency between stored velocity vectors and
speed magnitude, the prescribed pressure gauge, CFL ratios, normalized
velocity divergence, and flux imbalance. Normalized velocity divergence
and flux imbalance are subject to enforced rejection thresholds during
dataset validation. Pressure-reference validation further requires a nonempty
$\mathcal{P}_t$ for every output frame and a residual mean of the
gauge-normalized pressure over $\mathcal{P}_t$ no greater than
$10^{-4}$.

Dataset-integrity checks verify surface-array dimensions, surface-normal
lengths, pressure-unit consistency, and the finite-value policy. NaN
values are permitted for Eulerian velocity and pressure inside solid
cells and for near-surface samples whose interpolation stencils contain
no positive valid-fluid weight; values expected to represent valid
fluid observations must be finite. Publication requires an
interpolation-valid fraction of at least $95\%$ on average for each wind
condition and at least $90\%$ in every retained frame. Conditions that
fail these gates are excluded until the underlying issue is corrected
and the affected outputs are regenerated and revalidated.

A robot version is published only when all $16$ configured wind
conditions have successful execution records, complete $200$-frame raw
dense-field outputs, and the required $8192$-point tracked near-surface
data with passing quality checks. Partial simulations and incompatible
pipeline versions are excluded. Benchmark sequences are constructed
only after raw-run validation using the fixed zero-based temporal
selection described above.

Versioned simulation directories retain source references, mesh and
motion assets, simulation configurations, audit records, and artifact
hashes alongside the generated HDF5 outputs. The HDF5 files contain the
aligned volumetric and near-surface fields together with motion
transforms and metadata, while the accompanying audit and quality-control
records document the corresponding validation results.

\begin{figure*}[!t]
    \centering
    \includegraphics[width=\textwidth]{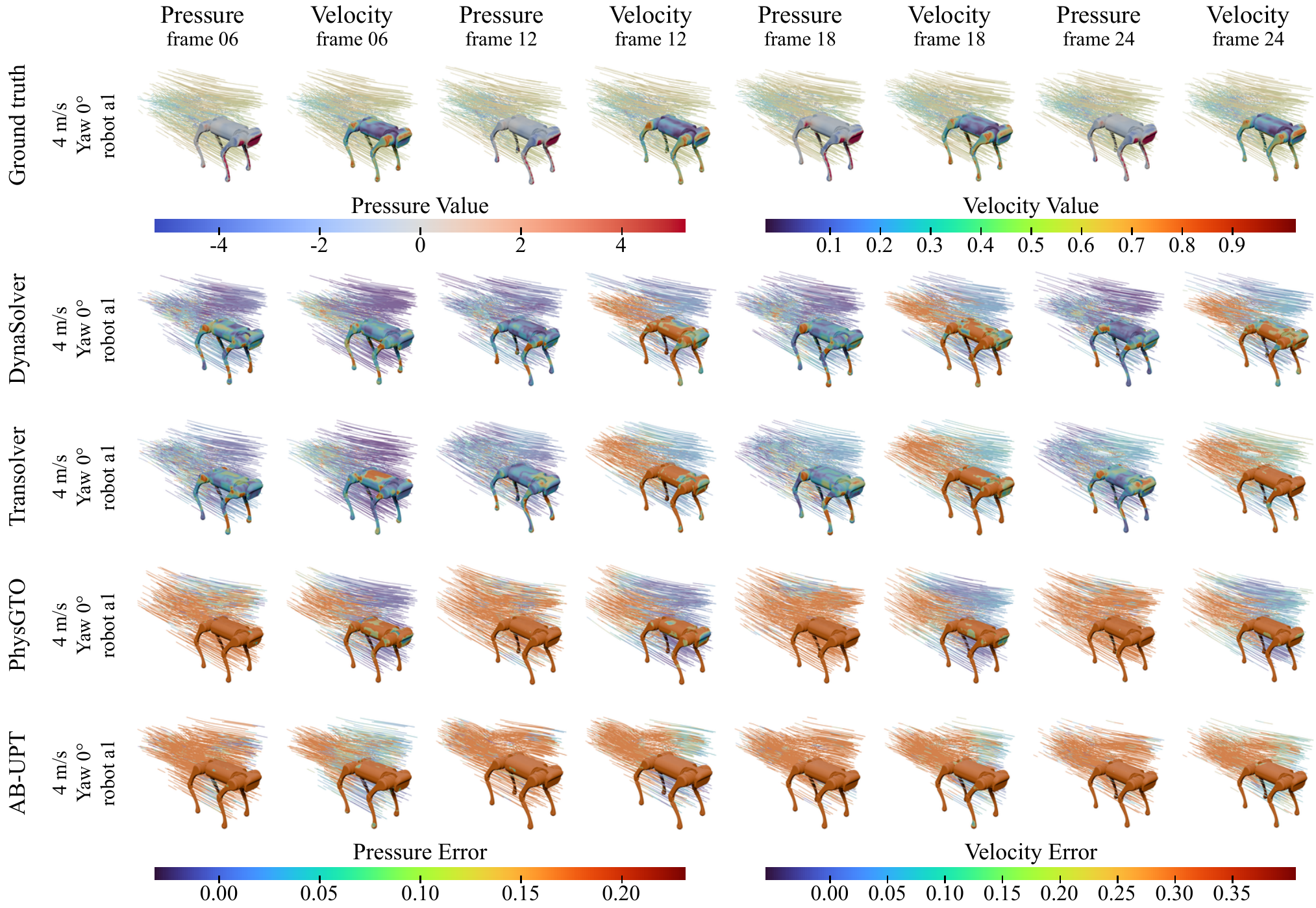}
    \caption{\textbf{Seen morphology under an unseen inflow condition.} Error maps of four models on a quadruped, evaluated at $4$~m/s and $0^\circ$ yaw. The morphology belongs to a family that the model sees during training, whereas this inflow condition is held out, so the figure isolates the effect of the flow regime with the boundary geometry in distribution. Columns alternate between the pressure and the velocity field at rollout steps 6, 12, 18 and 24, and all panels share one colour scale.
    }
\label{fig:qual_a1}
\end{figure*}

\begin{figure*}[!h]
    \centering
    \includegraphics[width=\textwidth]{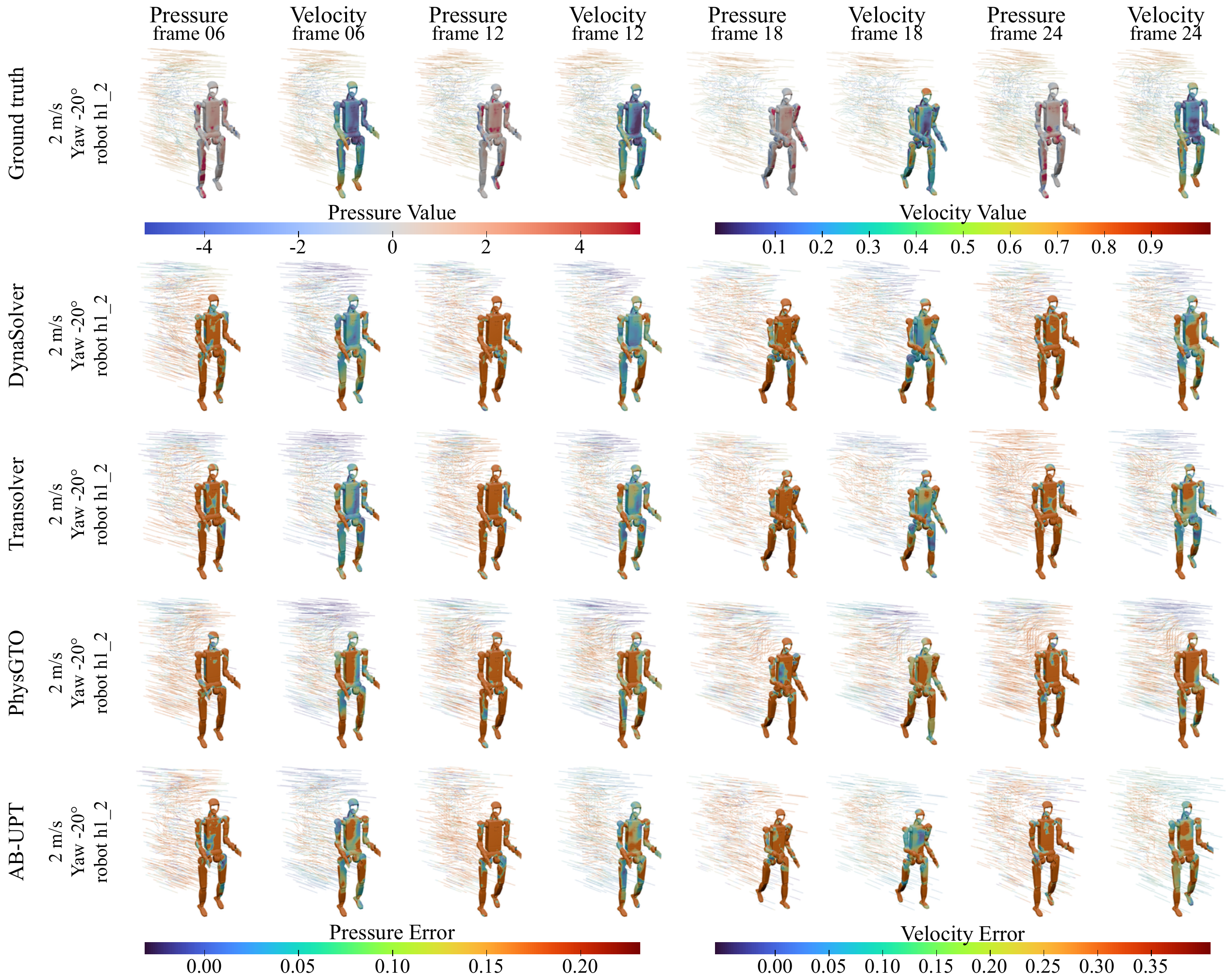}
    \caption{\textbf{Held-out morphology under an in-range inflow condition.}
The same layout as Figure~\ref{fig:qual_a1}, for an additional humanoid body
that is held out from training in all conditions, evaluated at $2$~m/s and
$-20^\circ$ yaw, which lies within the range of conditions the model was
trained on. The inflow therefore remains in distribution and the figure
isolates the effect of an unfamiliar morphology. Colour is shared by all
panels of the figure. 
}
\label{fig:qual_h1}
\end{figure*}

\section{Additional Qualitative Results}
We first visualize representative ground-truth trajectories from RoboCFD-4D in Fig.~\ref{fig:gt_01} and Fig.~\ref{fig:gt_02}. These examples show how the articulated boundary and the surrounding velocity and pressure fields evolve over time across different robot morphologies. They provide a direct view of the prediction targets, independent of any learned model.

The qualitative figure in the main text reports a humanoid that is held out from training. This appendix adds two cases that isolate the two axes of generalization separately. Figure~\ref{fig:qual_a1} keeps the morphology in distribution and changes the inflow: the quadruped belongs to a family that the model does see during training, but the case at $4$~m/s and $0^\circ$ yaw is held out, so a change in accuracy here is attributable to the flow regime rather than to an unfamiliar boundary. Figure~\ref{fig:qual_h1} does the opposite: it holds out an entire humanoid body, which does not appear in training under any condition, and evaluates it within the range of inflow conditions seen during training, so a change in accuracy here is attributable to the geometry. 

\begin{figure*}[!t]
    \centering
    \includegraphics[width=\textwidth]{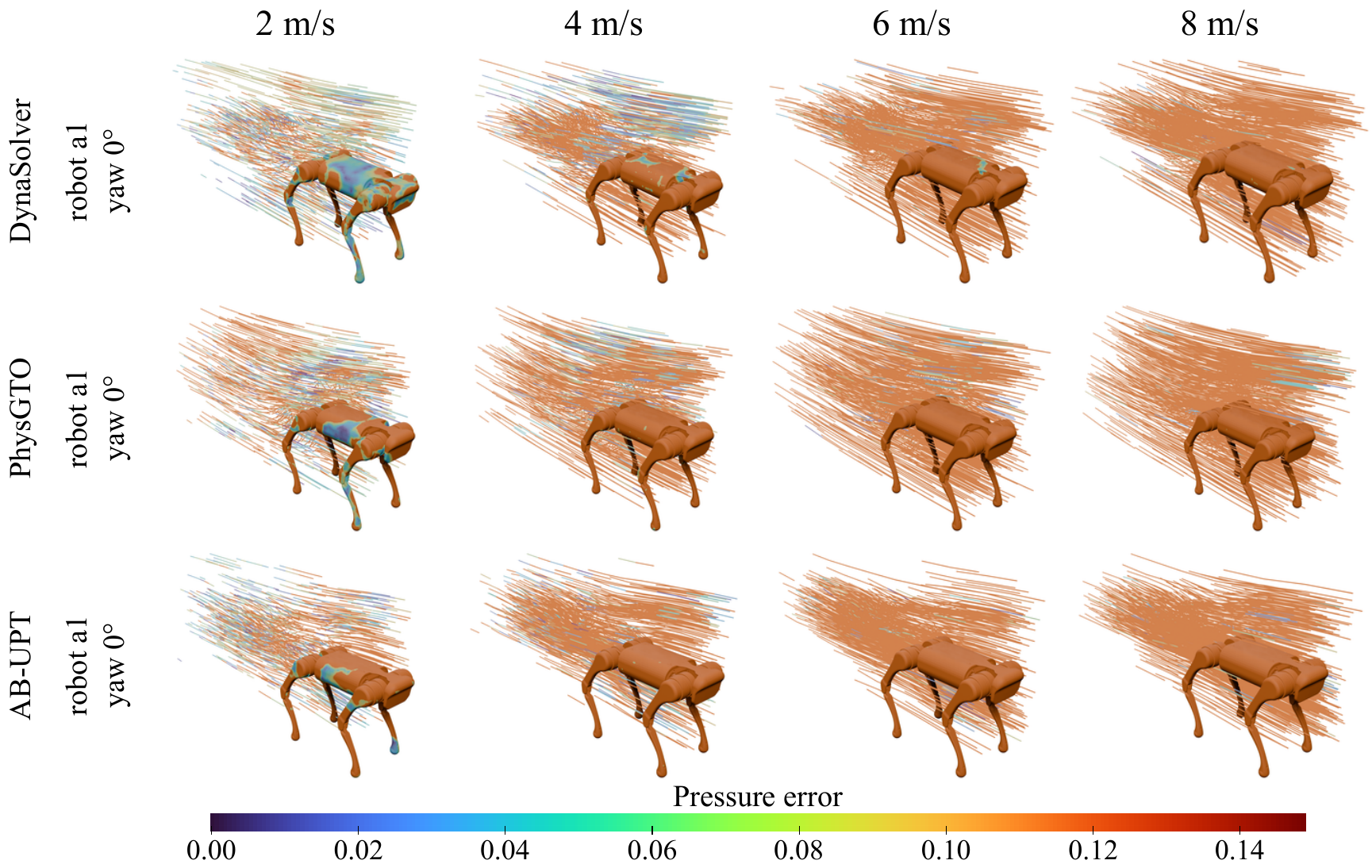}
    \caption{\textbf{Failure under stronger inflow conditions.}
Absolute pressure errors $|p_{\text{pred}}-p_{\text{gt}}|$ at rollout step $t=24$ for a quadruped at $0^\circ$ yaw under inflow speeds of $2$, $4$, $6$, and $8\,\mathrm{m/s}$. A shared color scale is used across all methods and conditions. While errors are relatively localized at low inflow speeds, they increase sharply and spread throughout the wake at higher speeds, indicating that strongly unsteady flow conditions remain challenging for current neural CFD surrogates.}
    \label{fig:high_wind_failure}
\end{figure*}

\section{Limitations and Future Work}
\label{sec:limitations}

\textbf{Benchmark Scope and Motion Realism.} RoboCFD-4D is entirely simulation-derived, with articulated motions generated from prescribed kinematic trajectories rather than dynamically simulated locomotion. Although the benchmark covers multiple inflow speeds and yaw angles, it does not capture the full diversity of realistic robotic motion. Existing real-robot datasets, such as CEAR, provide synchronized sensor measurements and joint encoder data across diverse quadrupedal gaits~\cite{zhu2024cear}. Building on such data, future work could incorporate measured or dynamically simulated locomotion trajectories to generate CFD sequences with more realistic boundary motion, enabling a more comprehensive evaluation of DynaSolver beyond the current prescribed-motion setting.

\textbf{Generalization to Unseen Geometries and Flow Regimes.} DynaSolver remains sensitive to changes in geometry and flow conditions. As shown in Fig.~\ref{fig:high_wind_failure}, pressure errors increase markedly with inflow speed and spread throughout the wake at $6$--$8\,\mathrm{m/s}$, highlighting the challenges of prediction under increasingly demanding flow conditions. Generalization to unseen robot geometries and the accumulation of errors over long autoregressive rollouts remain additional challenges. Related work on geometry-aware operator learning has also identified generalization to unseen shapes and error accumulation in long-horizon wake prediction as important challenges~\cite{rabeh2026predicting}. Future work could investigate improved geometry-conditioned representations and training strategies to enhance cross-morphology generalization and rollout stability under challenging flow conditions.